\documentclass[preprint,a4paper]{elsarticle}

\usepackage{graphicx}
\usepackage{amsmath}
\usepackage{float}
\usepackage{amsfonts}
\usepackage{a4wide}
\usepackage{algorithm}
\usepackage{algorithmicx}
\usepackage{algpseudocode}
\usepackage{epstopdf}
\usepackage{bm}
\usepackage{tikz}
\usepackage{multirow}
\usetikzlibrary{bayesnet}
\usetikzlibrary{shapes.gates.logic.US,trees,positioning,arrows}
\usepackage{caption}
\usepackage{subcaption}
\usetikzlibrary{trees}
\usepackage{graphicx}
\usetikzlibrary{	trees}
\usepackage{mathtools}
\usepackage{amsmath}
\usepackage{booktabs}
\usepackage[title,page]{appendix}

\usepackage{amsthm}

\usepackage{epstopdf}
\usepackage{amssymb}
\usepackage{microtype}
\usepackage{url}
\usepackage{caption}
\usepackage{subcaption}

\usepackage{xparse}

\journal{Mechanical Systems and Signal Processing}

\begin{document}
	
	\begin{frontmatter}
		
				\title{On risk-based active learning for structural health monitoring}

		\author{A.J.\ Hughes\corref{mycorrespondingauthor}}
		\cortext[mycorrespondingauthor]{Corresponding author}
		\ead{ajhughes2@sheffield.ac.uk}
		
		\author{L.A.\ Bull}
		
		\author{P.\ Gardner}
		
		\author{R.J.\ Barthorpe}
		
		\author{N.\ Dervilis}
		
		\author{K.\ Worden}
		
				\address{Dynamics Research Group, Department of Mechanical Engineering, University of Sheffield, \\ Sheffield, S1 3JD, UK}

		\begin{abstract}
A primary motivation for the development and implementation of structural health monitoring systems, is the prospect of gaining the ability to make informed decisions regarding the operation and maintenance of structures and infrastructure. Unfortunately, descriptive labels for measured data corresponding to health-state information for the structure of interest are seldom available prior to the implementation of a monitoring system. This issue limits the applicability of the traditional supervised and unsupervised approaches to machine learning in the development of statistical classifiers for decision-supporting SHM systems.

The current paper presents a risk-based formulation of active learning, in which the querying of class-label information is guided by the expected value of said information for each incipient data point. When applied to structural health monitoring, the querying of class labels can be mapped onto the inspection of a structure of interest in order to determine its health state. In the current paper, the risk-based active learning process is explained and visualised via a representative numerical example and subsequently applied to the Z24 Bridge benchmark. The results of the case studies indicate that a decision-maker's performance can be improved via the risk-based active learning of a statistical classifier, such that the decision process itself is taken into account.
		\end{abstract}
		
		\begin{keyword}
structural health monitoring \sep decision-making \sep active learning \sep value of information
		\end{keyword}
		
	\end{frontmatter}
	
	
\section{Introduction}

The field of structural health monitoring (SHM) is concerned with the development and implementation of online data acquisition and processing systems for the purpose of damage detection in structures and infrastructure \cite{Farrar2013}. Impelling advancement in SHM is the desire for quantitative decision support regarding the operation and maintenance (O\&M) of high-value and/or safety-critical assets. By the incorporation of information provided by SHM systems into the decision process, thereby facilitating condition-based O\&M, it is hoped that both structural safety can be improved and operational costs can be reduced.

One approach to decision-making in SHM is to adopt a probabilistic risk-based framework \cite{Hughes2021}, in which failure events and decidable actions - like maintenance - are assigned costs and utilities, respectively. Decisions are made so as to maximise expected utility gain or minimise expected utility loss. As a product of probability and utility, expected utility can be considered a measure of \textit{risk}~\cite{Bedford2001}. The risk-based framework employs probabilistic graphical modelling to form Bayesian network representations of fault trees to define a probability of a failure event conditioned on the health state of the structure. In addition, the approach uses transition/degradation models to forecast future health states. In accordance with \cite{Farrar2013}, the framework in \cite{Hughes2021} utilises a statistical pattern recognition (SPR) approach to damage detection and localisation in the inference of structural health states.

A critical challenge associated with the learning of statistical classifiers for SHM is a lack of labelled data corresponding to health-states of interest. This issue prevents the use of supervised learning. While unsupervised methods may be used to find statistical patterns within data, such patterns are of limited use in decision-support because of the lack of contextual information that would be provided by data labels. Several methods have been investigated as a means to overcome this challenge, including the use of physics-based models \cite{Barthorpe2021} and transfer learning \cite{Gardner2021b,Michau2019}. An alternative approach, that enables the online development of classifiers, is \textit{active learning}. An active learning framework for SHM has been developed in \cite{Bull2019}, in which probabilistic classifiers direct the acquisition of new labelled data according to an uncertainty measure, given the current model. Within the active learning framework for SHM, newly-acquired labelled data correspond to diagnostic information provided by an engineer, following an inspection of a structure. While this formulation provides a principled methodology for allocating inspection resource in a manner that optimises classification accuracy, in some scenarios it may be desirable/utility-optimal to consider the active learning of a classifier with respect to the context in which the classifier is being applied; supporting O\&M decision-making.

Active learning has seen limited use in health and performance monitoring applications. A study applying active learning methods to a generative model for predicting machining tool-wear is presented in \cite{Bull2019conf}. Artificial neural networks with active sampling have been utilised for image classification task to detect defects in civil structures \cite{Feng2017}. In \cite{Arellano2019}, a Bayesian convolutional neural network incorporating entropy-based active sampling is proposed as an approach for monitoring tools. Additionally, a particle filter-based damage progression model is aided by actively selected data in \cite{Chakraborty2015}. Previously, applications of active learning to SHM have all adopted an information-theoretic perspective. The current paper aims to formulate the active learning process from a decision-theoretic perspective. This goal is achieved by applying active learning in the context of probabilistic risk-based SHM and considering the \textit{expected value of perfect information} (EVPI) with respect to a maintenance decision process.

The layout of the paper is as follows. Sections \ref{sec:RBSHM} and \ref{sec:learning} provide background information on risk-based SHM and on machine learning for SHM, respectively. Section \ref{sec:RBAL} presents the methodology for conducting risk-based active learning. Section \ref{sec:example} demonstrates risk-based active learning of a probabilistic mixture model for a simple, but representative, numerical dataset. Section \ref{sec:Z24} presents the risk-based active approach to learning as applied to the Z24 Bridge benchmark dataset. Finally, Sections \ref{sec:discussion} and \ref{sec:conclusions} provide discussions and conclusions, respectively.

\section{Probabilistic Risk-Based SHM}
\label{sec:RBSHM}

\subsection{Probabilistic graphical models}

Probabilistic graphical models (PGMs) are graphical representations of factorisations of joint probability distributions and are a powerful tool for reasoning and decision-making under uncertainty. For this reason, they are apt for representing and solving decision problems in the context of SHM, where there is uncertainty in the health states of structures. While there exist multiple forms of probabilistic graphical model, the key types utilised for the risk-based decision frameworks are Bayesian networks (BNs) and influence diagrams (IDs) \cite{Sucar2015}.

Bayesian networks are directed acyclic graphs (DAGs) comprised of nodes and edges. Nodes represent random variables and edges connecting nodes represent conditional dependencies between variables. In the case where the random variables in a BN are discrete, the model is defined by a set of conditional probability tables (CPTs). For continuous random variables, the model is defined by a set of conditional probability density functions (CPDFs).

\begin{figure}[ht!]
	\centering
	\begin{tikzpicture}[x=1.7cm,y=1.8cm]
	

	\node[obs] (parent) {$X$} ;
	\node[latent, right=1cm of parent] (inter) {$Y$} ;
	\node[latent, right=1cm of inter] (child) {$Z$} ;
	
	\edge {parent} {inter} ; %
	\edge {inter} {child} ; %
	
	\end{tikzpicture}
	\caption{An example Bayesian network representing a factorisation of a joint probability distribution over three random variables $X$, $Y$ and $Z$.}
	\label{fig:PGM0}
\end{figure}
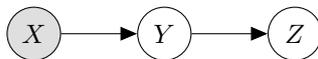

Figure \ref{fig:PGM0} shows a simple Bayesian network comprised of three random variables $X$, $Y$ and $Z$. $Y$ is conditionally dependent on $X$ and is said to be a \textit{child} of $X$, while $X$ is said to be a \textit{parent} of $Y$. $Z$ is conditionally dependent on $Y$ and can be said to be a child of $Y$ and a \textit{descendant} of $X$, while $X$ is said to be an \textit{ancestor} of $Z$. The factorisation described by the Bayesian network shown in Figure \ref{fig:PGM0} is given by $P(X,Y,Z) = P(X)\cdot P(Y|X)\cdot P(Z|Y)$. Given observations on a subset of nodes in a BN, inference algorithms can be applied to compute posterior distributions over the remaining unobserved variables. Observations of random variables are denoted in a BN via grey shading of the corresponding nodes, as is demonstrated for $X$ in Figure \ref{fig:PGM0}.

\begin{figure}[ht!]
	\centering
	\begin{tikzpicture}[x=1.7cm,y=1.8cm]
	

	\node[latent] (condition) {$W_c$} ;
	\node[obs,left=1cm of condition] (forecast) {$W_f$} ;
	\node[rectangle,draw=black,minimum width=0.7cm,minimum height=0.7cm,below=1cm of forecast] (decision) {$D$} ;
	\node[det, below=1cm of condition] (utility) {$U$} ;
	
	\edge {condition} {forecast} ; %
	\edge {forecast} {decision} ; %
	\edge {condition} {utility} ; %
	\edge {decision} {utility} ; %
	
	\end{tikzpicture}
	\caption{An example influence diagram representing the decision $D$ of whether to go outside or stay in under uncertainty in the future weather condition $W_c$ given an observed forecast $W_f$. Preferences for the various combinations of actions and weather conditions are specified by $U$.}
	\label{fig:ID0}
\end{figure}
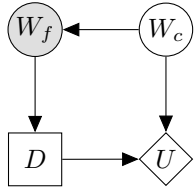

Bayesian networks may be adapted into influence diagrams to model decision problems. This augmentation involves the introduction of two additional types of node that are shown in Figure \ref{fig:ID0}: decision nodes, denoted as squares, and utility nodes, denoted as rhombi. For influence diagrams, edges connecting random variables to utility nodes denote that the utility function is dependent on the states of the random variables. Similarly, edges connecting decisions nodes to utility nodes denote that the utility function is dependent on the decided actions. Edges from decision nodes to random variable nodes indicate that the random variables are conditionally dependent on the decided actions. Edges from random variable or decision nodes to other decision nodes do not imply a functional dependence but rather order, i.e.\ that the observations/decisions must be made prior to the next decision being made.

To gain further understanding of IDs, one can consider Figure \ref{fig:ID0}. Figure \ref{fig:ID0} shows the ID for a simple binary decision; stay home and watch TV or go out for a walk, i.e.\ $domain(D) = \{ \text{TV},\text{ walk} \} $. Here, the agent tasked with making the decision has access to a weather forecast $W_{f}$ which is conditionally dependent on the future weather condition $W_{c}$. The weather forecast and future condition share the same possible states $domain(W_f) = domain(W_c) = \{ \text{bad}, \text{ good} \} $. The utility achieved $U(W_c,D)$, is then dependent on both the future weather condition and the decided action. For example, one might expect high utility gain if the agent decides to go for a walk and the weather condition is good.

In general, a policy $\delta$ is a mapping from all possible observations to possible actions. The problem of inference in influence diagrams is to determine an optimal strategy $\bm{\Delta}^{\ast} = \{ \delta^{\ast}_{1},\ldots, \delta^{\ast}_{n} \}$ given a set of observations on random variables, where $\delta^{\ast}_{i}$ is the $i^{th}$ decision to be made in a strategy $\bm{\Delta}^{\ast}$ that yields the \textit{maximum expected utility} (MEU). For further details on the computation of the MEU for influence diagrams, the reader is directed to \cite{Kjaerulff2008} and \cite{Koller2009}. Defined as a product of probability and utility, the expected utility can be considered as a quantity corresponding to risk.

\subsection{Decision Framework}

\begin{figure}[h!]
	\centering
	\begin{tikzpicture}[x=1.7cm,y=1.8cm]
	
	\node[det] (uf1) {$U_{F_{t}}$} ;
	\node[latent,below=1cm of uf1] (failure) {$F_{\bm{S}}$} ;
	\node[const, below=1cm of failure] (temp) {$ $} ;
	\node[latent, right=0.5cm of temp] (bay2) {$h\bm{s}_{2}$} ;
	\node[latent, left=0.5cm of temp] (bay1) {$h\bm{s}_{1}$} ;
	\node[const, below=1.5cm of temp] (temp2) {$ $} ;
	\node[latent, right=0.5cm of temp2] (mem6) {$hc_{3}$} ;
	\node[latent, right=1.5cm of temp2] (mem2) {$hc_{4}$} ;
	\node[latent, left=1.5cm of temp2] (mem5) {$hc_{1}$} ;
	\node[latent, left=0.5cm of temp2] (mem1) {$hc_{2}$} ;
	\node[det, right=3.6cm of uf1] (uf2) {$U_{F_{t+1}}$} ;
	\node[rectangle,draw=black,dashed,minimum width=1cm,minimum height=1cm,below=1cm of uf2] (ft2) {$F^\prime_{t+1}$} ;
	\node[latent, below=2cm of temp2] (x1) {$\bm{H}_{t}$} ;
	\node[latent, right=4cm of x1] (x2) {$\bm{H}_{t+1}$} ;
	\node[obs, below=1cm of x1] (y1) {$\bm{\nu}_{t}$} ;
	\node[rectangle,draw=black,minimum width=1cm,minimum height=1cm,below=1cm of y1] (d1) {$d_{t}$} ;
	\node[det, below=1cm of d1] (u1) {$U_{d_{t}}$} ;
	
	\edge {failure} {uf1} ; %
	\edge {ft2} {uf2} ; %
	\edge {x2} {ft2} ; %
	\edge {x1} {x2} ; %
	\edge {x1} {y1} ; %
	\edge {d1} {x2} ; %
	\edge {d1} {u1} ; %
	\edge {bay1} {failure} ; %
	\edge {bay2} {failure} ; %
	\edge {mem1} {bay1} ; %
	\edge {mem5} {bay1} ; %
	\edge {mem2} {bay2} ; %
	\edge {mem6} {bay2} ; %
	\edge {x1} {mem1} ; %
	\edge {x1} {mem5} ; %
	\edge {x1} {mem2} ; %
	\edge {x1} {mem6} ; %
	\edge {y1} {d1} ; %
	
	\matrix [draw,below right=1cm] at (current bounding box.north east) {
  \node[latent, below=0cm,label=right:Global health state] {$\bm{H}$}; \\
  \node [latent, below=0.2cm,label=right:Component health state] {$hc$}; \\
  \node [latent, below=0.2cm,label=right:Substructure health state] {$h\bm{s}$}; \\
  \node [latent, below=0.2cm,label=right:Failure state] {$F_{\bm{S}}$}; \\
  \node [obs, below=0.2cm,label=right:Observed features] {$\bm{\nu}$}; \\
  \node [rectangle, below=0.2cm,draw=black,label=right:Decision] {$d$}; \\
  \node [det, below=0.2cm,label=right:Utility] {$U$}; \\
  };
	
	\end{tikzpicture}
	\caption{An influence diagram representing a partially-observable Markov decision process over one time-slice for determining the utility-optimal maintenance strategy for a simple structure comprised of four components. The fault-tree failure-mode model for time $t+1$ has been represented as the node $F^\prime_{t+1}$ for compactness.}
	\label{fig:OverallPGM1}
\end{figure}
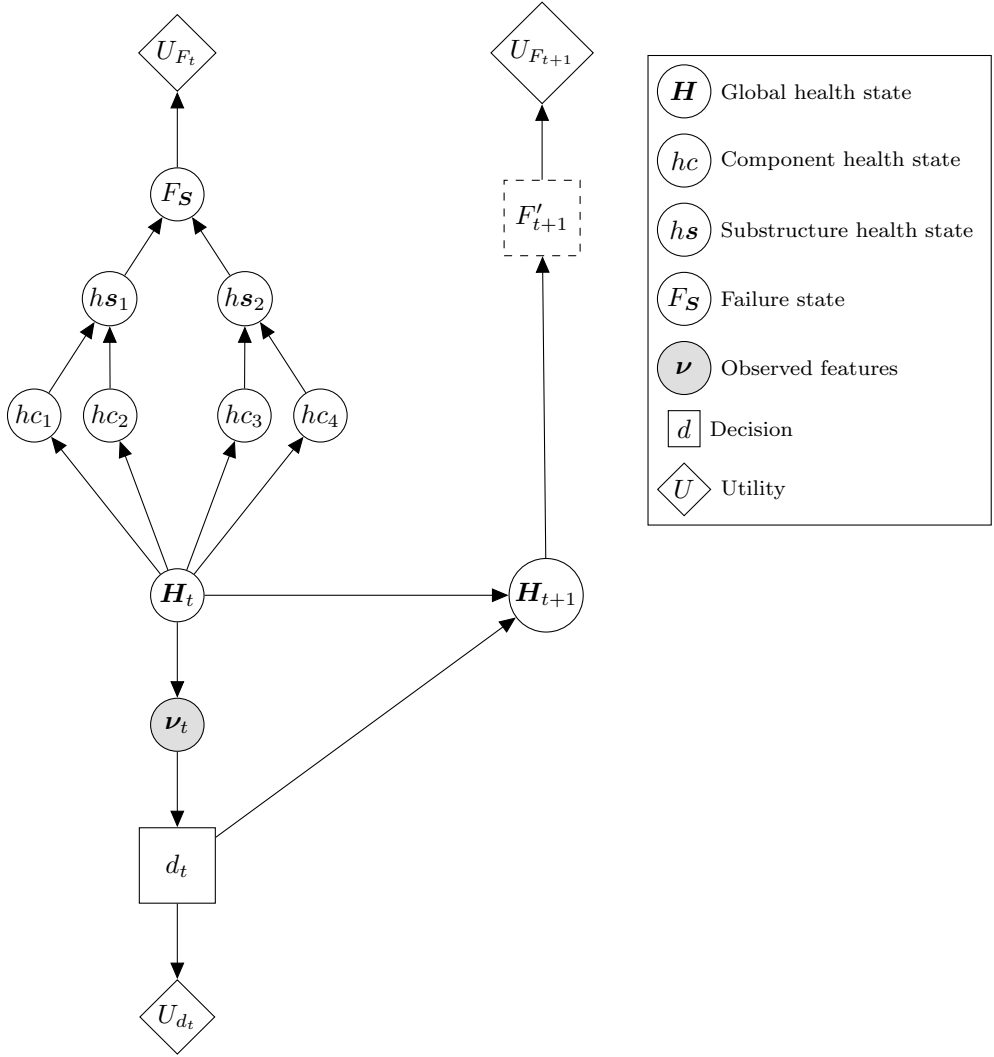

A probabilistic graphical model for a general SHM decision problem across a single time-slice is shown in Figure \ref{fig:OverallPGM1}. Here, a maintenance decision $d$ is shown for a simple fictitious structure $\bm{S}$, comprised of two substructures $\bm{s}_{1}$ and $\bm{s}_{2}$, each of which are comprised of two components; $c_{1,2}$ and $c_{3,4}$, respectively.

The overall decision process model shown in Figure \ref{fig:OverallPGM1} is based upon a combination of three sub-models; a statistical classifier, a failure-mode model, and a transition model.

Within the decision framework, a random variable denoted $\bm{H}_t$ is used to represent the latent global health state of the structure at time $t$. For this decision process, a posterior probability distribution over the latent health state $\bm{H}_t$ is inferred with a statistical classifiers via observations on a set of discriminative features $\bm{\nu}_t$. It is assumed that the generative conditional distribution $P(\bm{\nu}|\bm{H})$ is learned implicitly or explicitly, depending on the choice of statistical classifier. Here, the use of a probabilistic classifier is vital to ensure decisions made are robust to uncertainty in the health state of the structure.

The failure condition of the structure $F_{\bm{S}}$ is represented as a random variable within the PGM and is conditionally dependent on the health states of the substructures denoted by the nodes $h\bm{s}_1$ and $h\bm{s}_2$. The health states of the substructures are dependent on the local health states of the constituent components denoted by the nodes $hc_{1-4}$. The local health states of the components are summarised in the global health-state vector $\bm{H} = \{ hc_1, hc_2, hc_3, hc_4 \}$. The conditional probability tables defining the relationship between these random variables correspond to the Boolean truth tables for each of the logic gates in the fault tree defining the failure mode $F_{\bm{S}}$ \cite{Bobbio2001,Mahadevan2001}. This failure-mode model is repeated in each time-step. The failure states associated with the variable $F_{\bm{S}}$ are given utilities via the function represented by the node $U_F$. As it is necessary to consider the future risk of failure in the decision process, these utility functions are also repeated for each time-step.

Finally, a transition model is used to forecast the future health states given the current health state and a decided action, i.e.\ $P(\bm{H}_{t+1}|\bm{H}_{t},d_{t})$. The transition model considers the degradation of the structure under the various operational and environmental conditions a structure may experience, while accounting for uncertainties in each.

\section{Machine Learning Paradigms}
\label{sec:learning}

\subsection{Supervised and unsupervised learning for SHM}

In taking a data-driven statistical pattern recognition approach to SHM, one employs machine learning tools to learn patterns in data acquired from structures in order to infer information about structural health states such as the presence, location, and type of damage. For classification in general, the $i^{\text{th}}$ measured data point $\bm{x}_i \in X$ can be categorised according to a descriptive label $y_i \in Y$ where $y_i$ corresponds to the ground truth of the classification problem. For SHM, observations $\bm{x}_i$ correspond to features extracted from the raw data acquired from a structure via signal processing, and the descriptive labels $y_i$ relate to structural health-state information.

As aforementioned, probabilistic classifiers are desirable in SHM. For probabilistic classifiers, the features $\bm{x}_i$ are defined as random vectors existing in a $D$-dimensional feature space $X \in \mathbb{R}^D$. Additionally, the descriptive labels $y_i$ are defined by a discrete random variable such that $y_i \in Y = \{1,\ldots,K \}$ where $Y$ is the label space and $K$ is the number of classes required to uniquely identify the structural health states of interest.

Traditionally in SHM, classifiers are learned using one of two frameworks; \textit{supervised} or \textit{unsupervised} learning \cite{Murphy2012}.

For a supervised classifier $f$, a mapping between the feature space and the label space is learned, i.e.\ $f : X \rightarrow Y$. Supervised learning requires a fully-labelled training-set $\mathcal{D}_l$ such that \cite{Schwenker2014},

\begin{equation}
\mathcal{D}_l = \{ (\bm{x}_i , y_i) | \bm{x}_i \in X , y_i \in Y \}_{i=1}^{n}
\end{equation}

\noindent
for $n$ collected data points. In the context of SHM, a fully-labelled training-set is often prohibitively expensive to obtain, or otherwise unavailable.

Conversely, unsupervised learning techniques (e.g.\ $k$-means clustering \cite{Likas2003}) may be applied when only unlabelled data are available and the training-set $\mathcal{D}_u$ is of the form,

\begin{equation}
\mathcal{D}_u = \{ \bm{x}_i | \bm{x}_i \in X \}_{i=1}^{m}
\end{equation}

\noindent
for $m$ collected data points. The issue with unsupervised techniques is that, without label information corresponding to the structural health conditions, the models learned are of limited usefulness for decision-making. This drawback arises as there is no context associated with the model and thus a related decision process cannot be specified. For a more in-depth discussion of the use of supervised and unsupervised learning in SHM, the reader is directed to \cite{Bull2019}.

\subsection{Active learning for SHM}

Active learning is a form of \textit{partially-supervised learning} \cite{Schwenker2014}. Partially-supervised learning algorithms are characterised by their use of both labelled and unlabelled data, such that the dataset is,

\begin{equation}
\mathcal{D} = \mathcal{D}_l \cup \mathcal{D}_u
\end{equation}

\begin{figure}[pt]
	\centering
	\begin{tikzpicture}[auto]
	\begin{footnotesize}
	\tikzstyle{block} = [rectangle, thick, draw=black!80, text width=5em, text centered, minimum height=9em, fill=black!5]
	\tikzstyle{line} = [draw, -latex, thick]
	\node [block, node distance=24mm] (A) {Provide\\ unlabelled input data};
	\node [block, right of=A, node distance=24mm] (B) {Establish which data are the most informative};
	\node [block, right of=B, node distance=24mm] (C) {Provide labels for these data};
	\node [block, right of=C, node distance=24mm] (D) {Train a classifier on this informed subset};
	\path [line, draw=black!80] (A) -- (B);
	\path [line, draw=black!80] (B) -- (C);
	\path [line, draw=black!80] (C) -- (D);
	\end{footnotesize}
	\end{tikzpicture}
	\caption{The general active learning heuristic from \cite{Bull2019}.}
	\label{fig:AL1}
\end{figure}
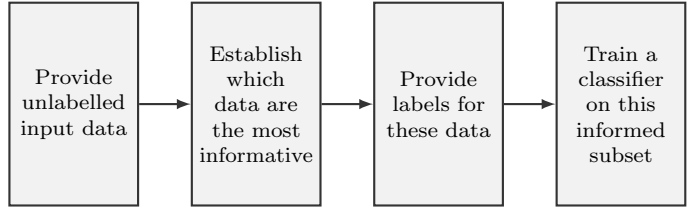

Active learning algorithms automatically query unlabelled data in $\mathcal{D}_u$ to obtain labels allowing the labelled dataset $\mathcal{D}_l$ to be extended. A generalised active learning heuristic is presented in Figure \ref{fig:AL1}.

The probabilistic active learning framework for SHM developed in \cite{Bull2019} details an approach built around a supervised probabilistic mixture model trained and retrained on $D_l$ as it is extended via the active querying process. The approach presented in \cite{Bull2019} uses measures of uncertainty to guide querying of incipient data points; specifically, preferentially obtaining labels for data points that have high entropy (information) \cite{MacKay2003} or low likelihood, given the current model. 

After some thought, one can realise that the active learning approach overcomes several of the challenges associated with supervised and unsupervised learning in SHM, as decision-making may be facilitated by the acquisition of class labels whilst limiting the expenditure necessary to obtain them.

\section{Risk-based Active Learning}
\label{sec:RBAL}

\begin{figure}[pt]
	\centering
	\begin{tikzpicture}[auto]
	\begin{footnotesize}
	\tikzstyle{block} = [rectangle, thick, draw=black!80, text width=5em, text centered, minimum height=9em, fill=black!5]
	\tikzstyle{line} = [draw, -latex, thick]
	\node [block, node distance=24mm] (A) {Provide\\ unlabelled input data};
	\node [block, right of=A, node distance=24mm] (B) {Establish which data have highest value of information};
	\node [block, right of=B, node distance=24mm] (C) {Provide labels for these data};
	\node [block, right of=C, node distance=24mm] (D) {Train a classifier on this informed subset};
	\path [line, draw=black!80] (A) -- (B);
	\path [line, draw=black!80] (B) -- (C);
	\path [line, draw=black!80] (C) -- (D);
	\end{footnotesize}
	\end{tikzpicture}
	\caption{The general risk-based active learning heuristic.}
	\label{fig:RBAL1}
\end{figure}
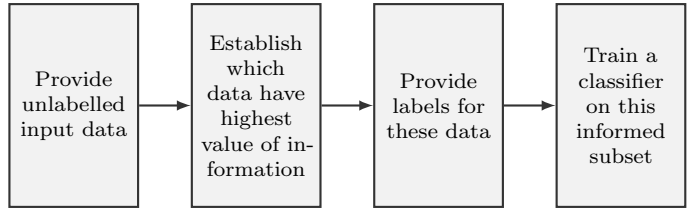

The current paper proposes a variation on the active learning framework presented in \cite{Bull2019} where, instead of using uncertainty measures to guide querying of data points according to their information or likelihood, incipient data are queried according to the expected value of perfect information with respect to the decision process, modelled using the risk-based SHM framework, in which the classifier is being applied. A generalised framework for risk-based active learning is presented in Figure \ref{fig:RBAL1}.

\subsection{Classifier initiation}

To begin the risk-based active learning process, one must initiate the classifier to be learned. Typically, this initialisation is done in a supervised manner with the available labelled data $\mathcal{D}_l$. In the case that there are no available labelled data, i.e.\ $\mathcal{D}_l = \emptyset$, and one has opted for a Bayesian learning approach, the classifier may be initiated using only the prior distribution.

Here, it is worth noting that the risk-based approach to active learning assumes that the number of classes (health states of interest) to be targeted by the model is known \textit{a priori}. Within an uncertainty-based approach to active learning, one can infer the number of classes from data \cite{Bull2019}. In contrast, for the risk-based approach to active learning, it is required that the classes targeted by the classifier correspond to those represented in the decision process. The prescription of the target classes limits the flexibility of the classifier, while also facilitating the computation of value of information.

\subsection{Value of information}

The expected value of perfect information (EVPI) is often understood as the price that a decision-maker should be willing to pay in order to gain access to perfect information regarding an otherwise uncertain or unknown state. More formally, EVPI can be defined as \cite{Koller2009},

\begin{equation}\label{eq:EVPI}
\text{EVPI}(d | X) := \text{MEU}(\mathcal{I}_{X \rightarrow d}) - \text{MEU}(\mathcal{I}) 
\end{equation}

\noindent
where $\text{EVPI}(d | X)$ is the expected value of observing (with perfect information) a variable $X$ before making a decision $d$, $\mathcal{I}$ corresponds to an original influence diagram for a decision process involving a decision node $d$ and a random variable node $X$, and $\mathcal{I}_{X \rightarrow d}$ corresponds to a modified influence diagram incorporating an additional edge from $X$ to $d$. The MEU for the decision process modelled by $\mathcal{I}$ is as follows,

\begin{equation}
	\text{MEU}(\mathcal{I}) = \max_{d} \sum_{y \in Y} P(y|d) \cdot U(y,d)
\end{equation}

\noindent where $Y$ is the subset of random variables in $\mathcal{I}$ with utility functions specified by $U$.

The EVPI has the important characteristic that it is strictly non-negative  (when disregarding the cost of making the additional observation), i.e.\ $\text{EVPI}(d | X) \ge 0$. This proposition can be realised by considering the conditional probability distributions over which the expected utility is optimised; the conditional probability distributions defined by the influence diagram $\mathcal{I}$ are a subset of those defined by $\mathcal{I}_{X \rightarrow d}$. Furthermore, $\text{EVPI}(d | X) = 0$ if and only if the optimal policy for $d$ in $\mathcal{I}$ remains optimal for $d$ in $\mathcal{I}_{X \rightarrow d}$; put simply, information has non-zero value when its possession results in a policy change.

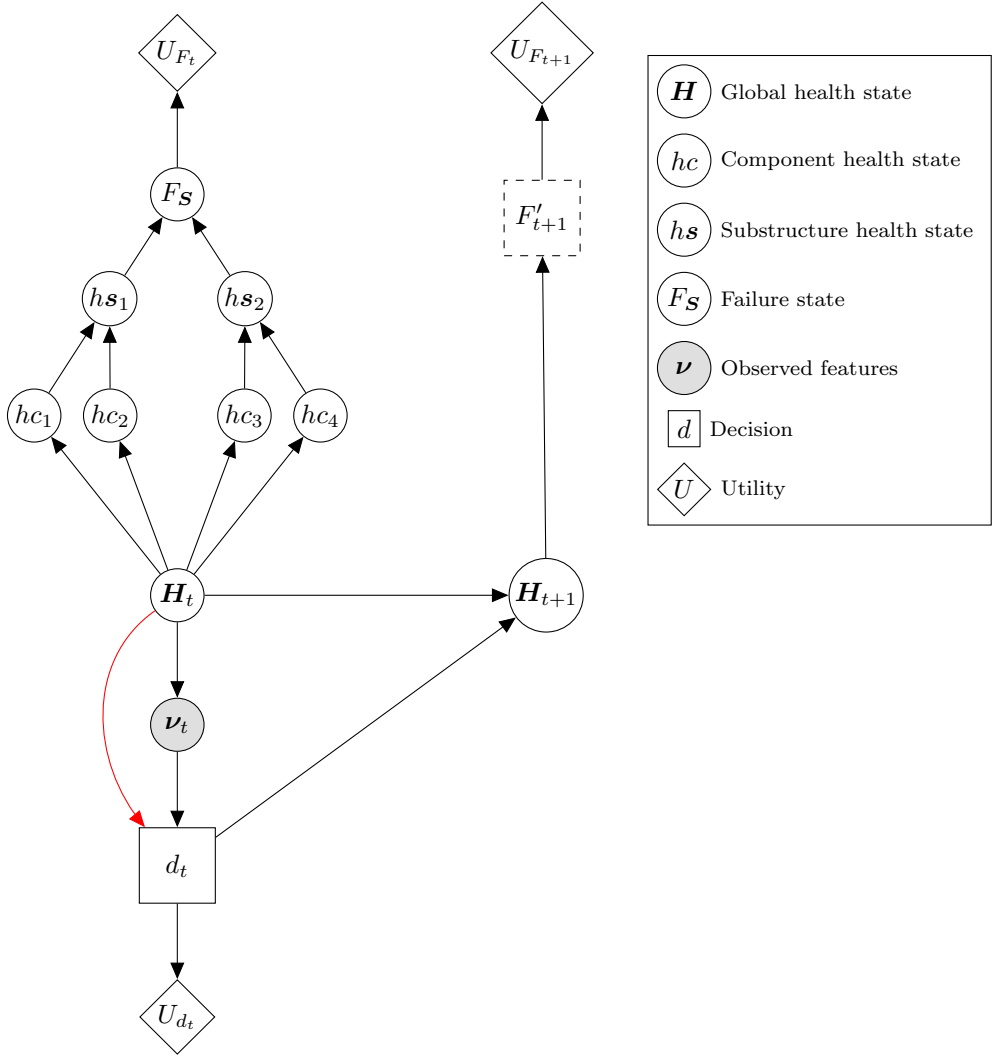
\begin{figure}[h!]
	\centering
	\begin{tikzpicture}[x=1.7cm,y=1.8cm]
	
	\node[det] (uf1) {$U_{F_{t}}$} ;
	\node[latent,below=1cm of uf1] (failure) {$F_{\bm{S}}$} ;
	\node[const, below=1cm of failure] (temp) {$ $} ;
	\node[latent, right=0.5cm of temp] (bay2) {$h\bm{s}_{2}$} ;
	\node[latent, left=0.5cm of temp] (bay1) {$h\bm{s}_{1}$} ;
	\node[const, below=1.5cm of temp] (temp2) {$ $} ;
	\node[latent, right=0.5cm of temp2] (mem6) {$hc_{3}$} ;
	\node[latent, right=1.5cm of temp2] (mem2) {$hc_{4}$} ;
	\node[latent, left=1.5cm of temp2] (mem5) {$hc_{1}$} ;
	\node[latent, left=0.5cm of temp2] (mem1) {$hc_{2}$} ;
	\node[det, right=3.6cm of uf1] (uf2) {$U_{F_{t+1}}$} ;
	\node[rectangle,draw=black,dashed,minimum width=1cm,minimum height=1cm,below=1cm of uf2] (ft2) {$F^\prime_{t+1}$} ;
	\node[latent, below=2cm of temp2] (x1) {$\bm{H}_{t}$} ;
	\node[latent, right=4cm of x1] (x2) {$\bm{H}_{t+1}$} ;
	\node[obs, below=1cm of x1] (y1) {$\bm{\nu}_{t}$} ;
	\node[rectangle,draw=black,minimum width=1cm,minimum height=1cm,below=1cm of y1] (d1) {$d_{t}$} ;
	\node[det, below=1cm of d1] (u1) {$U_{d_{t}}$} ;
	
	\edge {failure} {uf1} ; %
	\edge {ft2} {uf2} ; %
	\edge {x2} {ft2} ; %
	\edge {x1} {x2} ; %
	\edge {x1} {y1} ; %
	\edge {d1} {x2} ; %
	\edge {d1} {u1} ; %
	\edge {bay1} {failure} ; %
	\edge {bay2} {failure} ; %
	\edge {mem1} {bay1} ; %
	\edge {mem5} {bay1} ; %
	\edge {mem2} {bay2} ; %
	\edge {mem6} {bay2} ; %
	\edge {x1} {mem1} ; %
	\edge {x1} {mem5} ; %
	\edge {x1} {mem2} ; %
	\edge {x1} {mem6} ; %
	\edge {y1} {d1} ; %
	
	\draw [red, ->] (x1) to [out=-145,in=130] (d1);
	
	\matrix [draw,below right=1cm] at (current bounding box.north east) {
		\node[latent, below=0cm,label=right:Global health state] {$\bm{H}$}; \\
		\node [latent, below=0.2cm,label=right:Component health state] {$hc$}; \\
		\node [latent, below=0.2cm,label=right:Substructure health state] {$h\bm{s}$}; \\
		\node [latent, below=0.2cm,label=right:Failure state] {$F_{\bm{S}}$}; \\
		\node [obs, below=0.2cm,label=right:Observed features] {$\bm{\nu}$}; \\
		\node [rectangle, below=0.2cm,draw=black,label=right:Decision] {$d$}; \\
		\node [det, below=0.2cm,label=right:Utility] {$U$}; \\
		};
	
	\end{tikzpicture}
	\caption{A modified influence diagram $\mathcal{I}_{\bm{H}_t \rightarrow d_t}$. The additional edge is shown in red.}
	\label{fig:OverallPGM2}
\end{figure}

In the context of an SHM decision process, the expected value of inspection can be represented as $\text{EVPI}(d_t | \bm{H}_t)$, where the original influence diagram $\mathcal{I}$ corresponds to the decision process where the health state of the structure is inferred only via the observation of discriminative features with use of the statistical classifier. The modified influence diagram includes an additional edge from $\bm{H}_t$ to $d_t$ indicating the inspection of the structure. Considering the influence diagram shown in Figure \ref{fig:OverallPGM1} as $\mathcal{I}$, the modified influence diagram $\mathcal{I}_{\bm{H}_t \rightarrow d_t}$ is shown in Figure \ref{fig:OverallPGM2}. For the computation of the EVPI, it is assumed that inspection of the structure returns the ground-truth health state at the current time.

An example calculation of EVPI is provided in Section \ref{sec:EVPICalc}.

\subsection{Inspection scheduling}

\begin{figure*}[pt]
	\centering
	\scalebox{0.8}{
		\begin{tikzpicture}[auto]
		\begin{footnotesize}
		\tikzstyle{decision} = [diamond, draw, text width=4em, text badly centered, inner sep=2pt]
		\tikzstyle{block} = [rectangle, draw, text width=10em, text centered, rounded corners, minimum height=4em]
		\tikzstyle{block2} = [rectangle, draw, text width=10em, text centered, rounded corners, minimum height=4em, fill=black!5]
		\tikzstyle{line} = [draw, -latex']
		\tikzstyle{cloud} = [draw=black!50, circle, node distance=3cm, minimum height=2em, text width=4em,text centered]
		\tikzstyle{point}=[draw, circle]
		\node [block2, node distance=4em] (start) {start:\\ initial training-set, $\mathcal{D}_l$};
		\node [point, below of=start, node distance=12mm] (point) {};
		\node [block, below of=point, node distance=12mm] (train) {train model\\ $p(\bm{\nu}, \bm{H} | \mathcal{D}_l)$};
		\node [decision, below of=train, node distance=25mm] (new data) {new data?};
		\node [block2, left of=new data, node distance=40mm] (stop) {stop};
		\node [block, below of=new data, node distance=25mm] (update u) {update unlabelled set, $\mathcal{D}_u$};
		\node [cloud, left of=update u, node distance=40mm] (measured data) {new observation, $\bm{\nu}_t$};
		\node [block, below of=update u, node distance=25mm] (predict) {predict \\ $p(\bm{H}_t | \bm{\nu}_t,\mathcal{D}_l)$};
		\node [block, right of=predict, node distance=42mm] (EVPI) {compute \\ $\text{EVPI}(d_t | \bm{H}_t)$};
		\node [decision, above of=EVPI, node distance=25mm] (inspect) {$\text{EVPI} > C_{\text{ins}}$?};
		\node [block, above of=inspect, node distance=25mm] (query) {query health-state information for $\bm{\nu}_t$};
		\node [cloud, right of=query, node distance=40mm] (annotate) {$\bm{H}_t$ provided via inspection by engineer};
		\node [block, above of=query, node distance=25mm] (update l) {update $\mathcal{D}_l$ to include new queried labels};
		\path [line] (start) -- (point);
		\path [line] (point) -- (train);
		\path [line] (train) -- (new data);
		\path [line] (new data) -- node {no}(stop);
		\path [line] (new data) -- node {yes}(update u);
		\path [line, dashed, draw=black!50] (measured data) -- (update u);
		\path [line] (update u) -- (predict);
		\path [line] (predict) -- (EVPI);
		\path [line] (EVPI) -- (inspect);
		\path [line] (inspect) -- node {yes}(query);
		\path [line] (inspect) -- node {no}(new data);
		\path [line, dashed, draw=black!50] (annotate) -- (query);
		\path [line] (query) -- (update l);
		\path [line] (update l) |- (point);
		\end{footnotesize}
		\end{tikzpicture}
	}
	\caption{Flow chart to illustrate the risk-based active learning process for classifier development and inspection scheduling.}
	\label{fig:RBAL2}
\end{figure*}
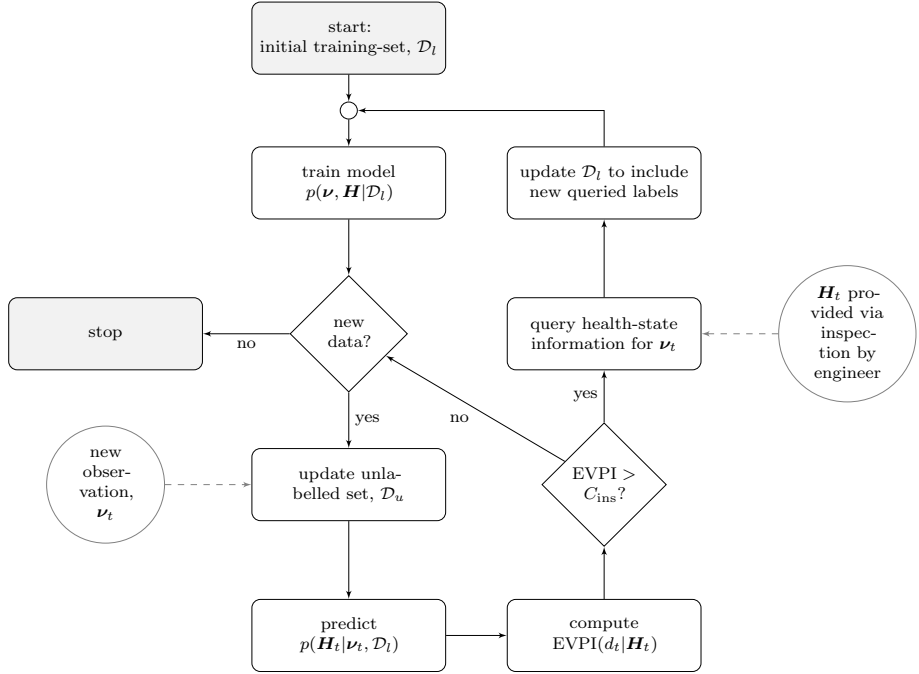

The EVPI of an unlabelled data point provides a convenient measure for determining whether a structure warrants inspection; if the EVPI for a data point $\bm{\nu}_t$ exceeds the cost of inspection $C_{\text{ins}}$, the structure should be inspected prior to $d_t$ and the corresponding health-state label for $\bm{H}_t$ obtained. Subsequently, $(\bm{\nu}_t,\bm{H}_t)$ can be incorporated into $\mathcal{D}_l$ and the classifier retrained. The risk-based active learning process for inspection scheduling and the development of statistical classifiers for risk-based SHM is shown in Figure \ref{fig:RBAL2}.

\subsection{Assessing performance}

Typically, classifier performance is evaluated using measures of classification accuracy, a popular choice being the $f_1$-score. As the focus of this paper is the development of classifiers in the context of decision-making, classification accuracy is of secondary concern. Rather here, an alternative metric for evaluating the more salient measure of decision-making performance is adopted, termed `decision accuracy' \cite{Hughes2021}.

Whereas, in the simplest sense, a classification accuracy is a comparison between the predicted outputs and the target outputs, `decision accuracy' is defined as a comparison between the actions selected by an agent using the statistical classifier being evaluated, and the optimal actions selected by an agent given perfect information, i.e. an agent in possession of the true target outputs of the classifier. A decision is considered `correct' if the agent utilising the classifier selects the same action as an agent with perfect information up to the current time. It follows that a decision can be considered incorrect if the decided action differs from that selected by an agent operating with perfect information. A quantitative value for decision accuracy may be calculated by taking the ratio:

\begin{equation}
\text{decision accuracy} = \frac{\text{number of correct decisions}}{\text{total number of decisions}}
\end{equation}

By virtue of its similarities with classification accuracy, decision accuracy provides a simple and intuitive metric to assess decision-making performance and is therefore deemed appropriate for the current paper. However, it is worth acknowledging here a limitation of decision accuracy as a measure of performance. In the same way that classification accuracy weights false-positives (type-I errors) and false-negatives (type-II errors) equally, decision accuracy considers the decision-making equivalents to be of equal concern. Naturally, in many SHM applications, one may prefer unnecessary action over neglectful inaction so as to avoid catastrophic structural failures. To account for such nuances, one may opt to use a utility-based metric \cite{Friedman2003}, or a weighted receiver operating characteristics (ROCs) \cite{Brown2006}. Additionally, these types of performance metric can also be extended to problems with non-binary decision domains and can reflect the relative preferability of candidate actions within such domains.

\section{Numerical Example}
\label{sec:example}

To demonstrate risk-based active learning for SHM in a visual manner, the framework was applied to a representative numerical case study. Consider a structure $S$ with four distinct health states of interest $H \in \{1, 2, 3, 4\}$:

\begin{itemize}
	\item State $1$ corresponds to the structure being undamaged and fully functional
	\item State $2$ corresponds to the structure possessing minor damage whilst being fully functional
	\item State $3$ corresponds to the structure possessing significant damage whilst operating at a reduced operational capacity
	\item State $4$ corresponds to the structure possessing critical damage and being non-operational, i.e.\ failed.
\end{itemize}

\subsection{Decision process}

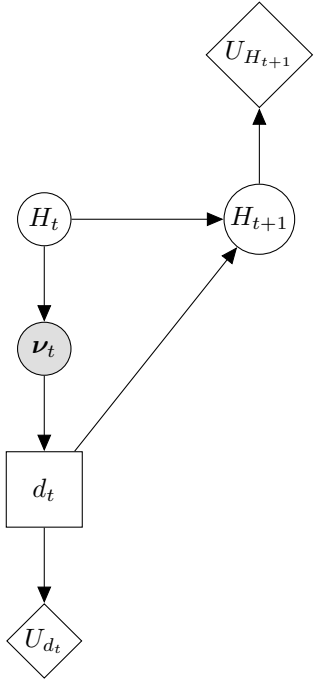
\begin{figure}[h!]
	\centering
	\begin{tikzpicture}[x=1.7cm,y=1.8cm]
	
	\node[det] (uf2) {$U_{H_{t+1}}$} ;
	\node[latent, below=1cm of uf2] (x2) {$H_{t+1}$} ;
	\node[latent, left=2cm of x2] (x1) {$H_{t}$} ;
	\node[obs, below=1cm of x1] (y1) {$\bm{\nu}_{t}$} ;
	\node[rectangle,draw=black,minimum width=1cm,minimum height=1cm,below=1cm of y1] (d1) {$d_{t}$} ;
	\node[det, below=1cm of d1] (u1) {$U_{d_{t}}$} ;

	\edge {x2} {uf2} ; %
	\edge {x1} {x2} ; %
	\edge {x1} {y1} ; %
	\edge {d1} {x2} ; %
	\edge {d1} {u1} ; %
	\edge {y1} {d1} ; %

	\end{tikzpicture}
	\caption{An influence diagram representation of the decision process associated with structure $S$.}
	\label{fig:NEPGM1}
\end{figure}

One can consider a decision process for the structure $S$, in which an agent at time $t$ is tasked with making a decision $d_t$, such that some degree of operational capacity is maintained and the structure is not in the failed state at time $t+1$. The influence diagram for such a decision process is shown in Figure \ref{fig:NEPGM1}.

Here, the decision $d_t$ is a binary choice where, $d_t = 0$ corresponds to `do nothing' and $d_t = 1$ corresponds to `perform maintenance'. The utilities associated with $d_t$ are specified by the utility function $U(d_t)$ and are shown in Table \ref{tab:Ud}. It is assumed that the `do nothing' action has no utility associated with it, whereas the `perform maintenance' action has negative utility relating to the expenditure associated with material and labour costs for structural maintenance. In practice, the specification of utility functions is non-trivial however may be achieved via expert elicitation during the operational evaluation stages of an SHM campaign. The elicitation of utility functions is beyond the scope of the current paper. Hence, for the purposes of the current case study, the relative values of utilities are selected to be somewhat representative of the SHM context.

\begin{table}[ht]
	\centering
	\caption{The utility function $U(d_t)$ where $d=0$ and $d=1$ denote the `do nothing' and `perform maintenance' actions, respectively.}
	\label{tab:Ud}   
	\begin{tabular}{c|c c}
		\hline\noalign{\smallskip}
		$d_t$ & $0$ & $1$\\
		\hline\noalign{\smallskip}
		$U(d_t)$ & $0$ & $-30$\\
		\noalign{\smallskip}\hline
	\end{tabular}
\end{table}

For the `do nothing' action $d_t = 0$, it is assumed that the structure monotonically degrades, with a propensity to remain in its current health state. This assumption is reflected in the CPD $P(H_{t+1}|H_t, d_t = 0)$ shown in Table \ref{tab:CPD1}.

\begin{table}[ht]
	\centering
	\caption{The conditional probability table $P(H_{t+1}|H_t, d_t)$ for $d_t = 0$.}
	\label{tab:CPD1}       
	\begin{tabular}{c c| c c c c}
		
		\multicolumn{2}{c|}{\multirow{4}{*}{ }} & \multicolumn{4}{c}{$H_{t+1}$} \\
		
		&  & 1 & 2 & 3 & 4 \\
		\hline\noalign{\smallskip}
		\multirow{4}{*}{$H_t$}   & 1 & 0.8 & 0.18 & 0.015 & 0.005\\
		& 2  & 0  & 0.8 & 0.15 & 0.05\\
		& 3  & 0  & 0 & 0.8 & 0.2\\
		& 4  & 0  & 0 & 0 & 1\\
		\hline
	\end{tabular}
\end{table}

For the `perform maintenance' action $d_t = 1$, it is assumed that the structure is returned to its undamaged health state with probability 0.99 and remains in its current state with probability 0.01. The conditional probability distribution $P(H_{t+1}|H_t, d_t = 0)$ is shown in Table \ref{tab:CPD2}.

\begin{table}[ht]
	\centering
	\caption{The conditional probability table $P(H_{t+1}|H_t, d_t)$ for $d_t = 1$.}
	\label{tab:CPD2}       
	\begin{tabular}{c c| c c c c}
		
		\multicolumn{2}{c|}{\multirow{4}{*}{ }} & \multicolumn{4}{c}{$H_{t+1}$} \\
		
		&  & 1 & 2 & 3 & 4 \\
		\hline\noalign{\smallskip}
		\multirow{4}{*}{$H_t$}   & 1 & 1 & 0 & 0 & 0\\
		& 2  & 0.99  & 0.01 & 0 & 0\\
		& 3  & 0.99  & 0 & 0.01 & 0\\
		& 4  & 0.99  & 0 & 0 & 0.01\\
		\hline
	\end{tabular}
\end{table}

For simplicity, within the decision process it is assumed that utilities may be attributed directly to the future health state $H_{t+1}$ of the structure without the need for modelling a specific failure mode. The utility function used for the current case study was specified so as to reflect the relative utility values that may be expected in a typical SHM application. The utility function $U(H_{t+1})$ is given in Table \ref{tab:UH}. Here, States 1 and 2, in which the structure $S$ is fully functional are assigned some positive utility, State 3 in which the structure is functional but with reduced capacity is assigned a lesser positive utility, and State 4 for which the structure is non-operational is assigned a relatively large negative utility to reflect the loss of functionality and some additional severe consequence associated with the failure, e.g.\ risk to human life. 

\begin{table}[ht]
	\centering
	\caption{The utility function $U(H_{t+1})$.}
	\label{tab:UH}       
	\begin{tabular}{c|c c c c}
		\hline\noalign{\smallskip}
		$H_{t+1}$ & $1$ & $2$ & $3$ & $4$\\
		\hline\noalign{\smallskip}
		$U(H_{t+1})$ & $10$ & $10$ & $5$ & $-75$\\
		\noalign{\smallskip}\hline
	\end{tabular}
\end{table}


Finally, it is also assumed that the health state $H_t$ may be inferred via the use of a statistical classifier by observing a set of discriminative features $\bm{\nu}_t = \{\nu_t^1,\nu_t^2 \}$. The ground-truth health state at time $t$ may be obtained via inspection at the cost of $C_{\text{ins}} = 7$.

\subsection{Statistical classifier}\label{sec:classifier}

While the risk-based active learning algorithm is not restricted to any particular type of classifier, the statistical model employed for the current case study is one similar to that used in \cite{Bull2019} - a mixture of four multivariate Gaussian distributions learned in a supervised manner from the initial labelled dataset $\mathcal{D}_l$. Each Gaussian component defines a generative model for the discriminative features $\bm{\nu}_t$, given each of four possible health states of interest in the domain of $H_t$,

\begin{equation}
p(\bm{\nu}_t | H_t = k) = \mathcal{N}(\bm{\mu}_k, \Sigma_k)
\end{equation}

\noindent
where $\bm{\mu}_k$ and $\Sigma_k$ are parameters of the multivariate Gaussian distribution corresponding to the mean and covariance, respectively. In addition to the mean and covariance parameters of the Gaussian components, the mixture model requires specification of $p(H_t)$,

\begin{equation}
H_t \sim \text{Cat}(\bm{\lambda})
\end{equation}

\noindent
where the categorical distribution is parametrised by a set of \textit{mixing proportions} $\bm{\lambda} = \{\lambda_1,\lambda_2,\lambda_3,\lambda_4\}$ such that,

\begin{equation}
P(H_t = k) = \lambda_k
\end{equation}

\noindent
and,

\begin{equation}
\sum_{k=1}^{4}P(H_t = k) = \sum_{k=1}^{4}\lambda_k = 1
\end{equation}

The parameters of the Gaussian mixture model that describe the generative statistical distribution $p(H_t, \bm{\nu}_t)$ can be summarised as,

\begin{equation}
\bm{\Theta} = \{(\bm{\mu}_1,\Sigma_1,\lambda_1),\ldots,(\bm{\mu}_4,\Sigma_4,\lambda_4)\}
\end{equation}

In order to learn the parameters $\bm{\Theta}$ from $\mathcal{D}_l$ in a manner that avoids over-fitting, a Bayesian approach was adopted here. In this approach, the parameters $\bm{\Theta}$ were treated as random variables with a prior placed over them. For conjugacy with the multivariate Gaussian distribution, a Normal-inverse-Wishart (NIW) prior was chosen such that,

\begin{equation}
\bm{\mu}_k , \Sigma_k \sim \text{NIW}(\bm{m}_0,\kappa_0,v_0,S_0)
\end{equation}

\noindent
where $\bm{m}_0$, $\kappa_0$, $v_0$ and $S_0$ are hyperparameters of the probabilistic mixture model. These hyperparameters can be interpreted in the following way \cite{Murphy2012}: $\bm{m}_0$ is the prior mean for each class mean $\bm{\mu}_k$, and $\kappa_0$ specifies the strength of the prior; $S_0$ is proportional to the prior mean for each class covariance $\Sigma_k$, and $v_0$ specifies the strength of that prior. The hyperparameters were specified such that each class $H_t$ was initially represented as a zero-mean and unit-variance Gaussian distribution.

As a conjugate to the categorical distribution, a Dirichlet prior was placed over the mixing proportions $\bm{\lambda}$,

\begin{equation}
\bm{\lambda} \sim \text{Dir}(\bm{\alpha})
\end{equation}

\noindent
and

\begin{equation}
p(\bm{\lambda}) \propto \prod_{k=1}^{4}\lambda_k^{\alpha_k-1}
\end{equation}

\noindent
where $\bm{\alpha} = \{\alpha_1,\ldots,\alpha_4\}$ are hyperparameters of the mixture model. The hyperparameters were specified such that the prior probability of each class in the mixture model was $\frac{1}{4}$, i.e.\ each state is equally weighted.

Posterior estimates of the parameters may be calculated using the labelled dataset $\mathcal{D}_l$. As conjugate priors were used, updates of the parameters may be computed analytically to obtain the posterior NIW distribution given by \cite{Gelman2013},

\begin{equation}
\bm{\mu}_k , \Sigma_k | H_t = k, \mathcal{D}_l \sim \text{NIW}(\bm{m}_n,\kappa_n,v_n,S_n)
\end{equation}

\noindent
where $\bm{m}_n$, $\kappa_n$, $v_n$, $S_n$ are the updated parameters and are computed as follows,

\begin{equation}
\bm{m}_n = \frac{\kappa_0}{\kappa_0 + n_k}\bm{m}_0 + \frac{n_k}{\kappa_0 + n_k}\bar{\bm{\nu}}_k
\end{equation}

\begin{equation}
\kappa_n = \kappa_0 + n_k
\end{equation}

\begin{equation}
v_n = v_0 + n_k
\end{equation}

\begin{equation}
S_n = S_0 + S + \kappa_0 \bm{m}_0 \bm{m}_0^{\top} - \kappa_n \bm{m}_n \bm{m}_n^{\top}
\end{equation}

\noindent
where $n_k$ is the number of observations in $\mathcal{D}_l$ with label $k$, $\bar{\bm{\nu}}_k$ is the sample mean of observations with label $k$, and $S$ is the empirical scatter matrix given by the uncentered sum-of-squares for observations in class $k$, $S = \sum_{i \in \mathbb{I}_k}\bm{\nu}_i\bm{\nu}_i^{\top}$ where $\mathbb{I}_k$ is the set of indices for observations with label $k$.

The posterior distribution of the mixing proportions $\bm{\lambda}$ remains Dirichlet, and is given by \cite{Gelman2013},

\begin{equation}
p(\bm{\lambda}|\mathcal{D}_l) \propto \prod_{k=1}^{4}\lambda_k^{n_k + \alpha_k-1}
\end{equation}

To make class predictions for unlabelled data in $\mathcal{D}_u$, the posterior predictive distributions over the labels and observations can be obtained by marginalising out the parameters of the model. The posterior predictive distribution for unlabelled observations is obtained via the following marginalisation,

\begin{equation}
p(\bm{\nu}_t|H_t = k, \mathcal{D}_l) = \int\int p(\bm{\nu}_t|\bm{\mu}_k,\Sigma_k) p(\bm{\mu}_k,\Sigma_k|H_t = k, \mathcal{D}_l)d\bm{\mu}_k d\Sigma_k
\end{equation}

\noindent
 resulting in the Student-\textit{t} distribution \cite{Murphy2012},

\begin{equation}
\bm{\nu}_t | H_t = k, \mathcal{D}_l \sim \newline \mathcal{T}(\bm{m}_n,\frac{\kappa_n + 1}{\kappa_n(v_n - D + 1)}S_n,v_n - D +1)
\end{equation}

\noindent
where $\bm{m}_n$, $\kappa_n$, $v_n$, $S_n$ are the updated hyperparameters and $D$ is the dimensionality of the feature space. Here, the first two parameters of the Student-\textit{t} distribution correspond to the mean and scale, respectively. The third parameter specifies the \textit{degrees of freedom}. The full functional form of the Student-\textit{t} distribution can be found in \cite{Murphy2012}.

Similarly, the posterior predictive distribution over the labels is obtained via the following marginalisation,

\begin{equation}
p(H_t|\mathcal{D}_l) = \int p(H_t| \bm{ \lambda}) p( \bm{\lambda} | \mathcal{D}_l) d \bm{ \lambda}
\end{equation}

\noindent
resulting in,
\begin{equation}
p(H_t = k | \mathcal{D}_l) = \frac{n_k + \alpha_k}{n + \alpha_0}
\end{equation}

\noindent
where $n = \sum_{k=1}^{4}n_k$ and $\alpha_0 = \sum_{k=1}^{4}\alpha_k$.

Finally, the predictive distribution for the class labels given a new unlabelled observation $\bm{\nu}_t$ can be obtained using the posterior predictive distribution and applying Bayes' rule \cite{Bull2019},

\begin{equation}
p(H_t = k|\bm{\nu}_t,\mathcal{D}_l) = \frac{p(\bm{\nu}_t|H_t = k, \mathcal{D}_l)p(H_t=k|\mathcal{D}_l)}{p(\bm{\nu}_t|\mathcal{D}_l)}
\end{equation}

In summary, a Gaussian mixture model was trained in a supervised Bayesian manner on $\mathcal{D}_l$ and subsequently retrained as $\mathcal{D}_l$ was extended via the risk-based active querying process. The classifier developed allows a probability distribution over possible health states to be obtained following an observation of the discriminative features.

\subsection{Example EVPI calculation}\label{sec:EVPICalc}

To further elucidate the risk-based active learning approach, an illustrative calculation of the EVPI is provided. This calculation is based upon the current case study as presented in Section \ref{sec:example}.
	
Consider Figure \ref{fig:NEPGM1} as the influence diagram of the original decision process $\mathcal{I}$. The corresponding modified influence diagram $\mathcal{I}_{\bm{H}_t \rightarrow d_t}$ is shown below in Figure \ref{fig:NEPGM2}.

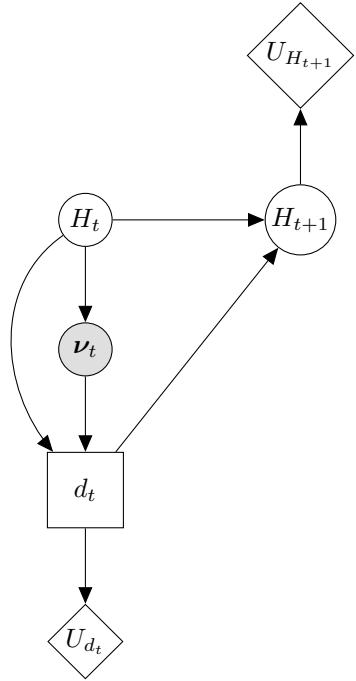
\begin{figure}[h!]
	\centering
	\begin{tikzpicture}[x=1.7cm,y=1.8cm]
	
	\node[det] (uf2) {$U_{H_{t+1}}$} ;
	\node[latent, below=1cm of uf2] (x2) {$H_{t+1}$} ;
	\node[latent, left=2cm of x2] (x1) {$H_{t}$} ;
	\node[obs, below=1cm of x1] (y1) {$\bm{\nu}_{t}$} ;
	\node[rectangle,draw=black,minimum width=1cm,minimum height=1cm,below=1cm of y1] (d1) {$d_{t}$} ;
	\node[det, below=1cm of d1] (u1) {$U_{d_{t}}$} ;

	\edge {x2} {uf2} ; %
	\edge {x1} {x2} ; %
	\edge {x1} {y1} ; %
	\edge {d1} {x2} ; %
	\edge {d1} {u1} ; %
	\edge {y1} {d1} ; %
	
	\draw [->] (x1) to [out=-145,in=130] (d1); %

	\end{tikzpicture}
	\caption{An influence diagram representation of the modified decision process associated with structure $S$.}
	\label{fig:NEPGM2}
\end{figure}

With the EVPI given by equation (\ref{eq:EVPI}), one must first calculate the MEU for the decision process represented by $\mathcal{I}$. The MEU for $\mathcal{I}$ is given by,

\begin{equation}\label{eq:MEU1}
\text{MEU}(\mathcal{I}) = \max_{d_t} \big[ \sum_{H_{t}}\sum_{H_{t+1}} P(H_t|\bm{\nu}_t) P(H_{t+1}|H_t,d_t) U(H_{t+1}) + U(d_t) \big]
\end{equation}

For the purposes of the demonstration, it will be assumed that the classifier presented in Section \ref{sec:classifier} has predicted $P(H_t|\bm{\nu}_t) = \{0.4, 0.3, 0.2, 0.1 \}$, otherwise, the remaining conditional probability distributions and utility functions are as specified in Section \ref{sec:example}. Employing equation (\ref{eq:MEU1}), one can compute that $\text{MEU}(\mathcal{I}) = -4.4$.

Next, one must calculate the MEU for the decision process represented by the influence diagram $\mathcal{I}_{H_t \rightarrow d_t}$. This is given by,

\begin{equation}\label{eq:MEU2}
\text{MEU}(\mathcal{I}_{H_t \rightarrow d_t}) = \sum_{H_{t}} \max_{d_t} \big[ \sum_{H_{t+1}} P(H_t|\bm{\nu}_t) P(H_{t+1}|H_t,d_t) U(H_{t+1}) + U(d_t) \big]
\end{equation}

Applying equation (\ref{eq:MEU2}) and again taking $P(H_t|\bm{\nu}_t) = \{0.4, 0.3, 0.2, 0.1 \}$, one determines $\text{MEU}(\mathcal{I}_{H_t \rightarrow d_t}) = 1.015$.

Finally, the EVPI of observing $H_t$ prior to making the decision $d_t$, or the expected value of inspecting the structure, can be trivially calculated  with equation (\ref{eq:EVPI}) to be,

\begin{equation}
\nonumber
\text{EVPI}(d_t|H_t) = 1.015 - -4.4 = 5.415
\end{equation} 

As per the risk-based active learning procedure outlined in Figure \ref{fig:RBAL2}, calculated values of EVPI, such as the one presented above, can be compared to $C_{\text{ins}}$ to determine whether an inspection is necessary.

\subsection{Results}

\begin{figure}[htbp!]
	\centering
	\scalebox{0.4}{
		\includegraphics{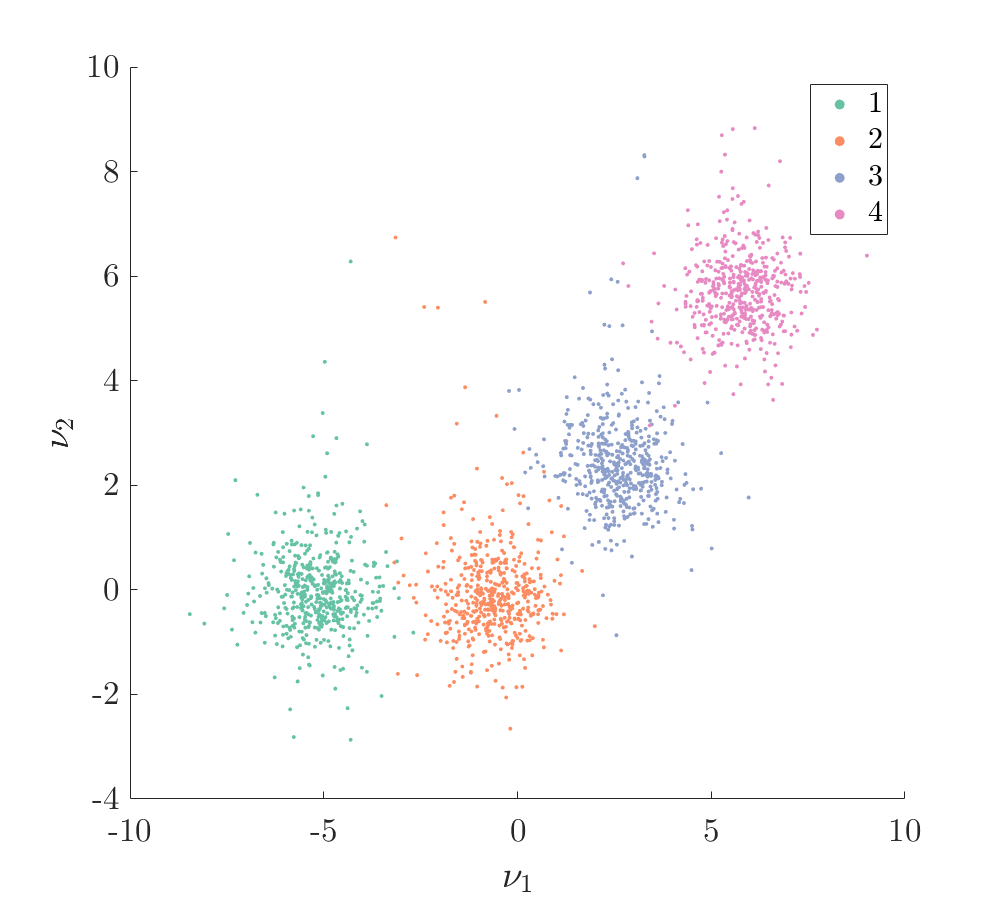}
	}
	\caption{The 2-dimensional feature space of the complete dataset.}
	\label{fig:AllData}
\end{figure}

The complete dataset associated with the structure $S$ in its various health states of interest $H_t \in \{1,2,3,4\}$ was comprised of 1997 data points and is shown in Figure \ref{fig:AllData}. The data were generated to be representative of typical changes in SHM feature spaces due to progressive damage. One half of the data were randomly selected and set aside to form an independent test set, the remaining data were used to form the dataset $\mathcal{D}$.

\begin{figure}[htbp!]
	\centering
	\scalebox{0.4}{
		\includegraphics{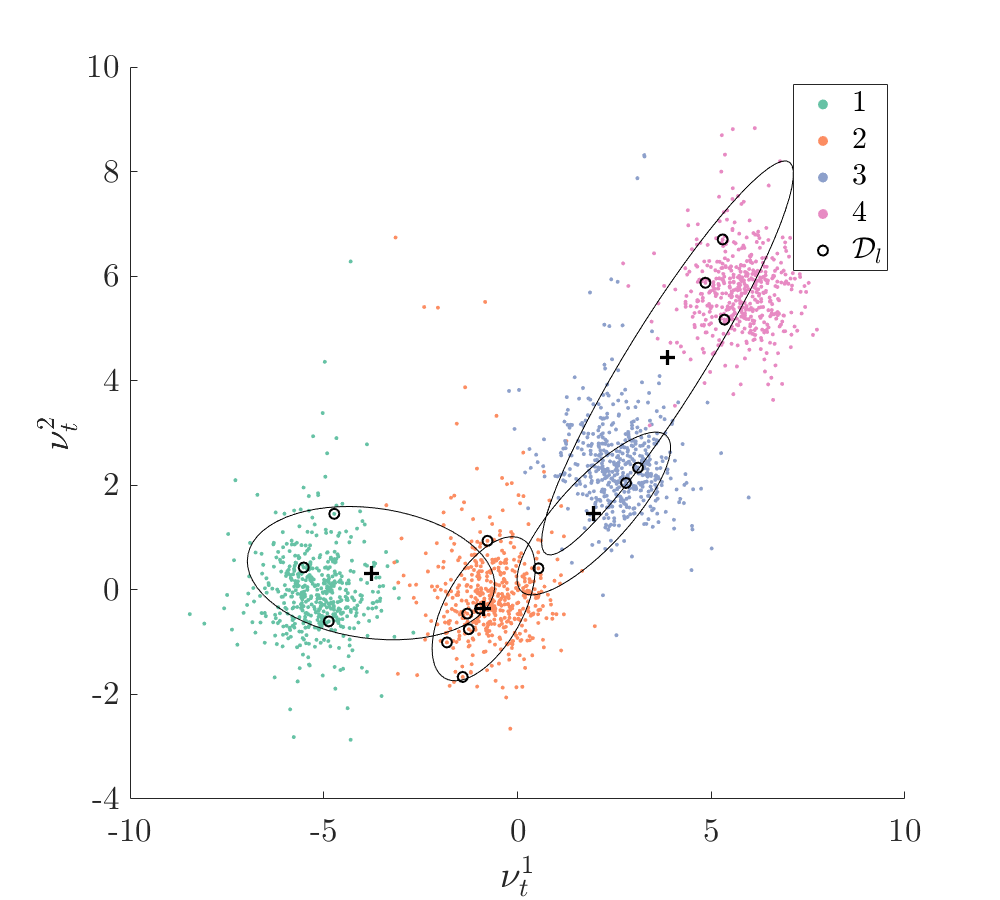}
	}
	\caption{An initial statistical classifier $p(\bm{\nu}_t, H_t, \bm{\Theta})$ learned from the original labelled dataset $\mathcal{D}_l$; \textit{maximum a posteriori} (MAP) estimate of the mean (\textbf{+}) and covariance (ellipses represent $2\sigma$).}
	\label{fig:InitialModel}
\end{figure}

A small (1.5\%) random subset of $\mathcal{D}$ was annotated with their corresponding labels to initialise $\mathcal{D}_l$ and the statistical classifier; it was ensured that at least one data point from each of the four classes was included in $\mathcal{D}_l$ upon initialisation. The remaining data $\mathcal{D}_u$ were left unlabelled to be sequentially presented to the decision model in the risk-based active learning process. An example of an initial model learned from $\mathcal{D}_l$ is shown in Figure \ref{fig:InitialModel} with data points in $\mathcal{D}_l$ circled.

\begin{figure}[htbp!]
	\centering
	\scalebox{0.4}{
		\includegraphics{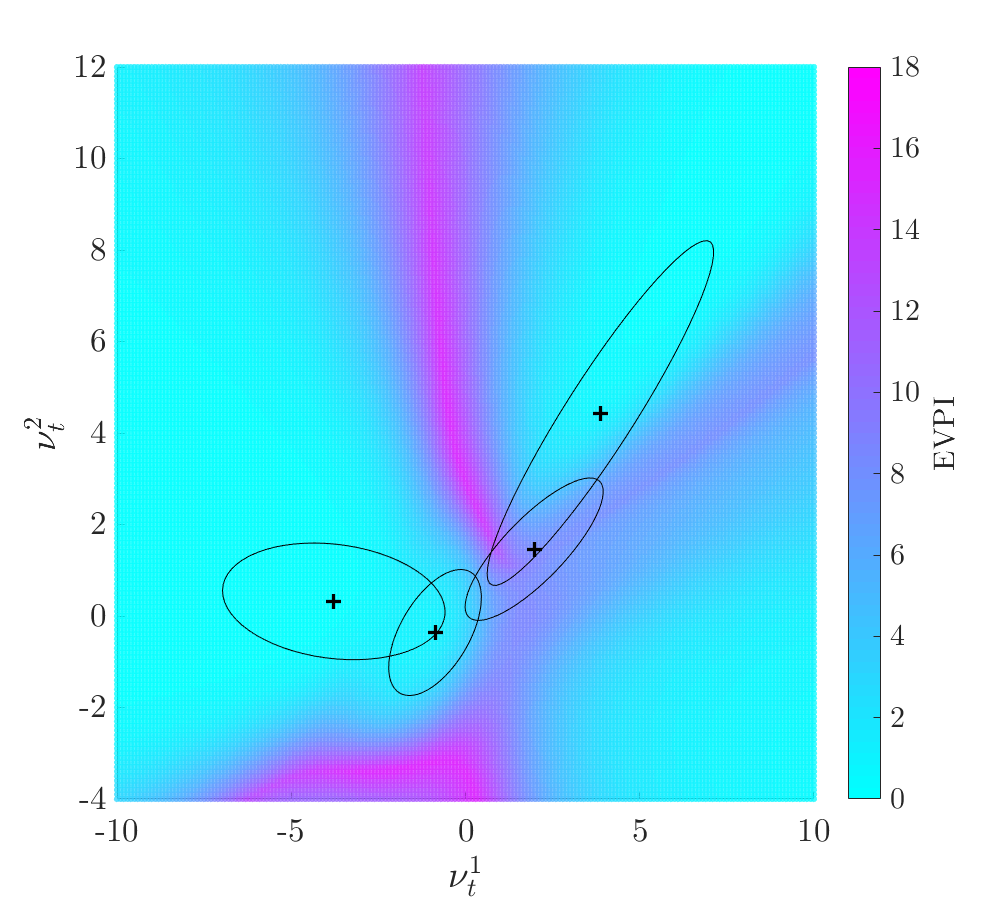}
	}
	\caption{The EVPI over the two-dimensional feature space given the initial model shown in Figure \ref{fig:InitialModel}.}
	\label{fig:InitialModelEVPI}
\end{figure}

It can be seen from Figure \ref{fig:InitialModel} that, in general, the initial model fits the data poorly as a result of insufficient data. The best fitted Gaussian component of the mixture model is for $H_t = 2$; this is to be expected, since this cluster lies close to the zero-mean of the prior distribution. The EVPI across the feature space given the initial model is shown in Figure \ref{fig:InitialModelEVPI}.

\newpage
Figure \ref{fig:InitialModelEVPI} shows the regions of high and low value of information. At this stage, it is worth reminding oneself that the expected value of information arises when obtaining information results in a change in policy; areas of high EVPI correspond to regions in the feature space where there is classification uncertainty in the vicinity of decision boundaries given the current model. Bearing this in mind, one can justify the observation that the feature space around the learned Gaussian component for Class 1 (corresponding to the structure being undamaged) has low EVPI. Additionally, the region of the feature space in the overlap between the learned Gaussian distributions for Class 1 and Class 2 (minor damage), where there is uncertainty between the two classes, has \textit{low} EVPI as the optimal policy is unchanged regardless of whether the structure is in State 1 or State 2. On the other hand, areas with high EVPI correspond to regions of the feature space where there is classification uncertainty between a benign state and a more worrisome state; an example of this is the dominant vertical band of high EVPI feature space that corresponds to the set of points equidistant (Mahalanobis distance) from the learned clusters for State 1 and State 4.

\begin{figure}[htbp!]
	\centering
	\scalebox{0.4}{
		\includegraphics{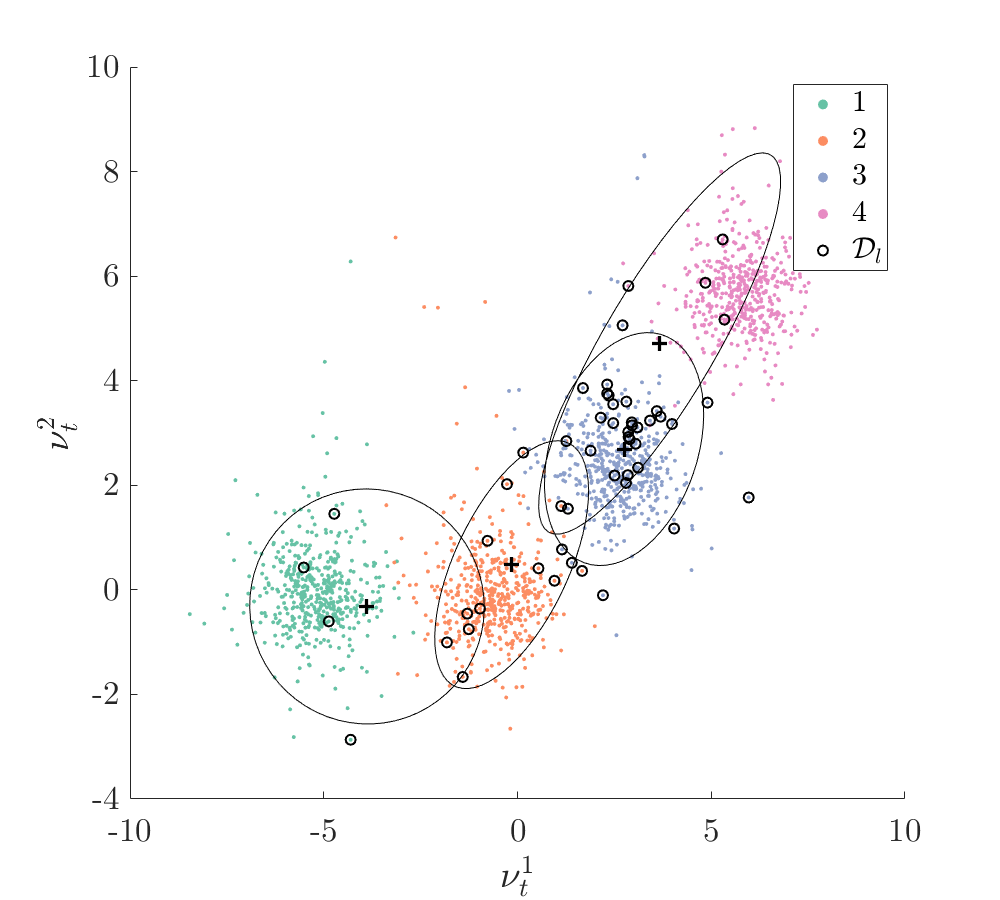}
	}
	\caption{A final statistical classifier $p(\bm{\nu}_t, H_t, \bm{\Theta})$ learned from the extended labelled dataset $\mathcal{D}_l$; \textit{maximum a posteriori} (MAP) estimate of the mean (\textbf{+}) and covariance (ellipses represent $2\sigma$).}
	\label{fig:FinalModel}
\end{figure}

The labelled dataset $\mathcal{D}_l$ was extended according to the risk-based active learning process presented in Section \ref{sec:RBAL}. Data points $\bm{\nu}_t$ in $\mathcal{D}_u$ were considered in random order one-at-a-time and had their EVPI computed given the current model. A data point was queried, annotated with a label, and incorporated into $\mathcal{D}_l$ if the data point met the condition $\text{EVPI} > C_{\text{ins}}$. After each query, the statistical classifier was retrained on the newly-extended $\mathcal{D}_l$. After being presented with each data point in $\mathcal{D}_u$ one at a time,
a final model trained on the fully-extended dataset in $\mathcal{D}_l$ is shown in Figure \ref{fig:FinalModel}.

It can be seen from Figure \ref{fig:FinalModel} that data in Classes 2 and 3 have been preferentially queried, particularly in the overlap between clusters; this is to be expected because of the associated classification uncertainty between states that warrant different maintenance policies. It can be seen that very few samples have been made for data points belonging to States 1 and 4, resulting in poorly-fitting learned distributions. This result can be explained by again considering the EVPI of the feature space shown in Figure \ref{fig:InitialModel}, which shows low value of information in the regions around those data.

\begin{figure}[htbp!]
	\centering
	\scalebox{0.4}{
		\includegraphics{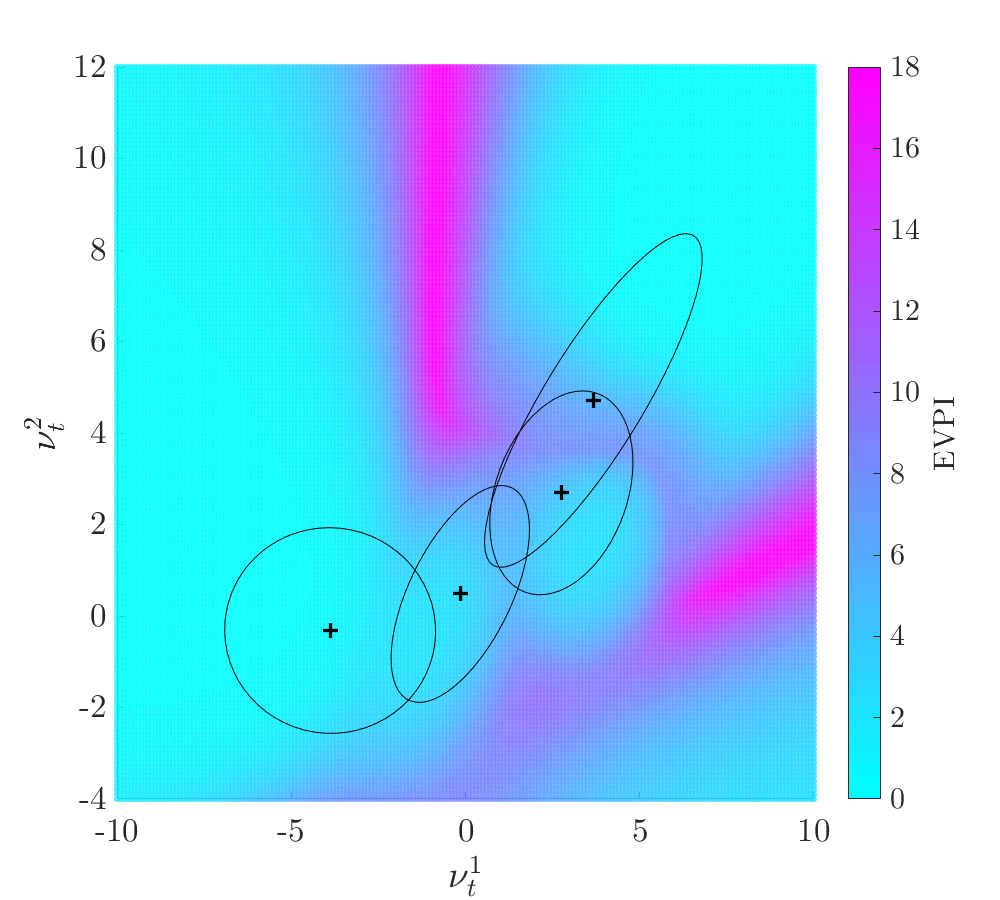}
	}
	\caption{The EVPI over the two-dimensional feature space given the final model shown in Figure \ref{fig:FinalModel}.}
	\label{fig:FinalModelEVPI}
\end{figure}

Figure \ref{fig:FinalModelEVPI} shows the EVPI over the feature space after the statistical classifier has been trained on the fully-extended version of $\mathcal{D}_l$. Similarities can be seen between Figure \ref{fig:InitialModelEVPI} and Figure \ref{fig:FinalModelEVPI} such as areas of low EVPI and the bands of high EVPI in regions equidistant from clusters corresponding to states that warrant differing maintenance actions. Figure \ref{fig:FinalModelEVPI} also shows a region of high EVPI that lies between the estimated means for the distributions learned for States 3 and 4; this is where the boundary between the structure being operational, and the structure being non-operational and requiring repair, lies. This result may indicate that the risk-based active learning approach preferentially incorporates data that strengthen decision boundaries.

\begin{figure}[htbp!]
	\centering
	\scalebox{0.4}{
		\includegraphics{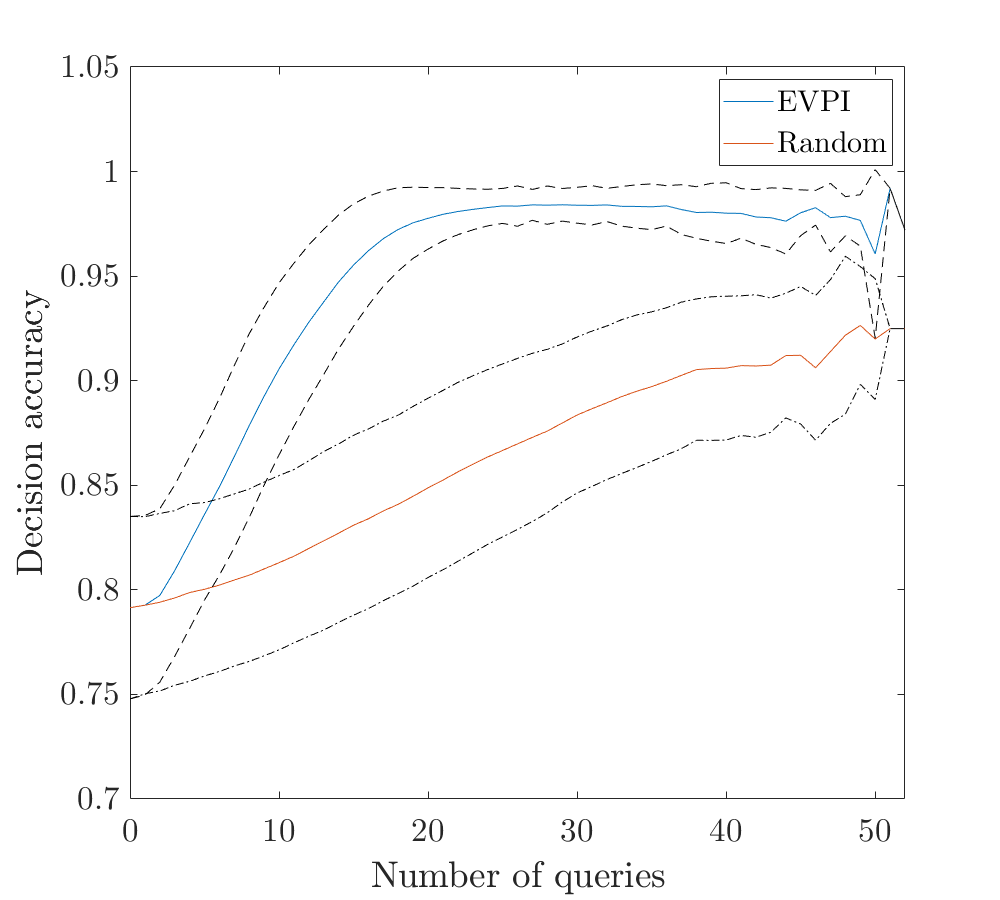}
	}
	\caption{The variation in decision accuracy with number of label queries for an agent utilising a statistical classifier trained on $\mathcal{D}_l$ extended via (i) risk-based active querying (EVPI) and (ii) random sampling (Random). The dashed lines show $\pm1\sigma$.}
	\label{fig:DecisionAccuracy}
\end{figure}

To evaluate the change in classifier decision performance throughout the active learning process, the decision accuracy achieved on the independent test set was computed after each retraining of the classifier. The active learning process was repeated 1000 times using different random number generator seeds so that the data selected to be in $\mathcal{D}$ and the initial subset $\mathcal{D}_l$ were randomly varied. The mean and standard deviation of the decision accuracy as a function of the number of queries is shown in Figure \ref{fig:DecisionAccuracy}. The mean and standard deviation of the decision accuracy for a classifier trained on $\mathcal{D}_l$ when extended with randomly-queried data points is also shown for comparison. 

The decision accuracy can be seen to increase more rapidly when guided by querying according to EVPI rather than random sampling, which suggests that incorporating data labels with high expected value into a classifier will improve an agent's decision-making. After approximately 20 queries, the improvement in decision accuracy gained per query is greatly reduced. Upon close examination, it can be seen that after 28 queries, the decision accuracy begins to decrease slightly. This may be a result of sampling bias introduced via the active learning process and will be further discussed in Section \ref{sec:discussion}.

In summary, a risk-based approach to active learning was demonstrated for a representative numerical case study. A simple maintenance decision problem was formed for a structure with four key health states of interest, and EVPI was used to trigger `inspections' to obtain class-label information corresponding to structural health states.

\section{Experimental Case Study}
\label{sec:Z24}

The risk-based active learning framework was applied to a dataset obtained from the Z24 Bridge \cite{Maeck2003}. The Z24 bridge was a concrete highway bridge near Solothurn in Switzerland, between the municipalities of Koppigen and Utzenstorf. Prior to demolition, the bridge was the focus of a cross-institutional research project (SIMCES), the goal of which was to provide a benchmark and prove the feasibility of vibration-based SHM \cite{DeRoeck2003,Maeck2001}. The Z24 Bridge benchmark has been widely used for SHM and modal analysis applications. The monitoring campaign on the bridge spanned a period of 12 months, during which time the bridge was instrumented with sensors to capture both the dynamic response of the structure and environmental conditions including air temperature, deck temperature, humidity and wind speed \cite{Peeters2001}. 

\begin{figure}[htbp!]
	\centering
	\scalebox{0.4}{
		\includegraphics{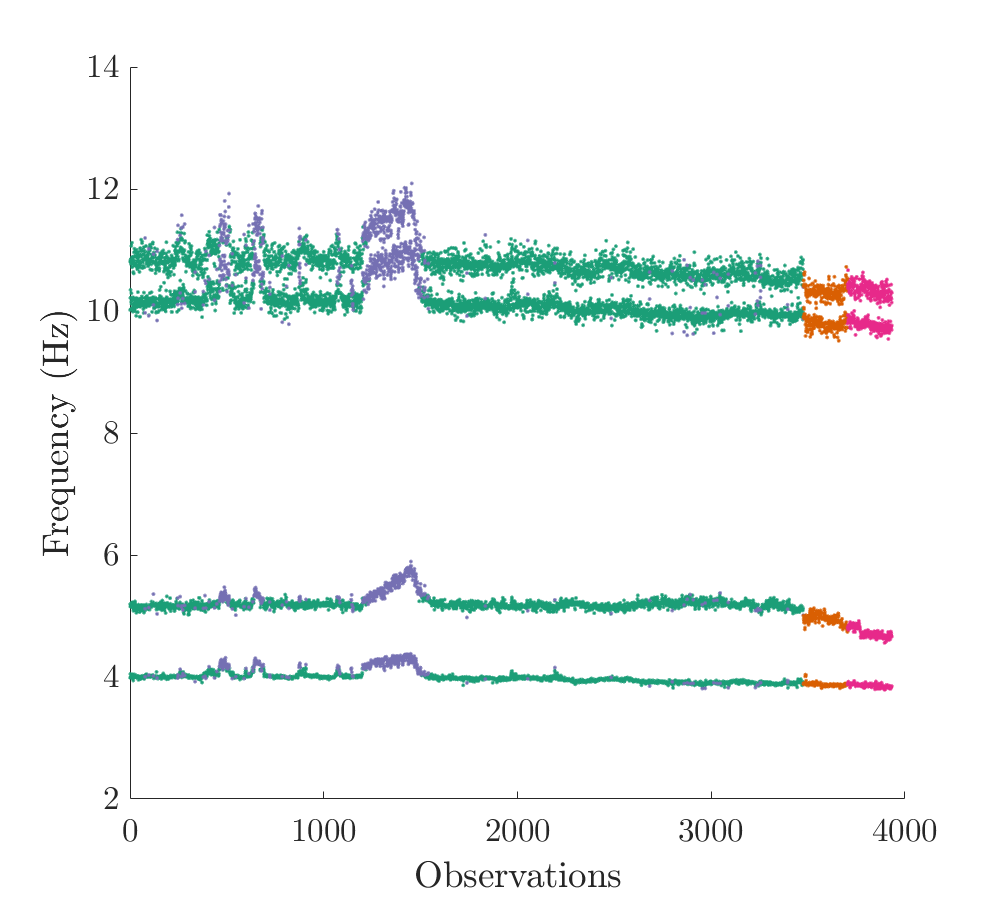}
	}
	\caption{Time history of the first four natural frequencies for the Z24 bridge.}
	\label{fig:z24overview}
\end{figure}

From the dynamic response measurements acquired, the first four natural frequencies of the structure were obtained. The variation in these natural frequencies throughout the monitoring campaign are shown in Figure \ref{fig:z24overview}. The dataset contains 3932 observations of the natural frequencies in total. Towards the end of the campaign, incremental damage was introduced to the structure artificially, beginning at observation 3476. Additionally, throughout the campaign, the bridge exhibited cold temperature effects, particularly prominent between observations 1200 and 1500. These are believed to be a result of the stiffening of the asphalt layer in the bridge deck induced by very low ambient temperatures.

To define a classification problem on which to apply risk-based active learning, the first four natural frequencies of the bridge were selected as the discriminative features such that $\bm{\nu}_t \in \mathbb{R}^4$. Furthermore, there are assumed to be four distinct classes of interest $H_t \in \{ 1,2,3,4 \}$:

\begin{itemize}
	\item Class 1: normal undamaged condition (green)
	\item Class 2: cold temperature undamaged condition (blue)
	\item Class 3: incipient damaged condition (orange)
	\item Class 4: advanced damaged condition (pink).
\end{itemize}

Here, it has been assumed that the damaged data may be separated into two halves; the earlier half corresponding to incipient damage and the later half corresponding to advanced damage. It is believed that this is a reasonable assumption, given the incremental addition of damage during the monitoring campaign \cite{Maeck2003}. Furthermore, it is supported by the unsupervised clustering achieved using the Dirichlet process in \cite{Rogers2019}. The normal undamaged data are separated from the cold temperature effects using outlier analysis via the Minimum Covariance Determinant algorithm \cite{Dervilis2014,Rousseeuw1999}.

\subsection{Decision process}

Again, to undertake risk-based active learning, one must consider the decision process for the structure. As the bridge is long since demolished, here, a similar decision process to that considered for the visual example presented in Section \ref{sec:example} will be considered. This decision process focusses on a binary decision $d_t$ where $d_t = 0$ corresponds to `do nothing' and $d_t = 1$ corresponds to `perform maintenance'. Because of the similar nature of the problems, the influence diagram shown in Figure \ref{fig:NEPGM1} can be used to represent the decision process considered for the Z24 bridge. However, the utility functions and conditional probability distributions defined by the influence diagram must be altered to reflect the operational context of the bridge.

The utilities associated with the decision $d_t$ specified by the utility function $U(d_t)$ are presented in Table \ref{tab:UdZ24}. It is assumed that the `do nothing' action has zero utility, whereas the `perform maintenance' action has negative utility. Once again, the utilities for the current case study have been selected such that the relative values are appropriate for demonstrating the risk-based active learning approach. As discussed earlier, more representative or exact utilities may be obtained with the aid of expert judgement. 

\begin{table}[ht]
	\centering
	\caption{The utility function $U(d_t)$ for the Z24 bridge where $d=0$ and $d=1$ denote the `do nothing' and `perform maintenance' actions, respectively.}
	\label{tab:UdZ24}   
	\begin{tabular}{c|c c}
		\hline\noalign{\smallskip}
		$d_t$ & $0$ & $1$\\
		\hline\noalign{\smallskip}
		$U(d_t)$ & $0$ & $-100$\\
		\noalign{\smallskip}\hline
	\end{tabular}
\end{table}

For the decision action `do nothing', the state transitions are specified such that the structure monotonically degrades, with a propensity to remain in its current state. For the current case study, however, there is the subtlety that States 1 and 2 both correspond to the undamaged health state but under differing environmental conditions. Here, the earlier assumption that there is a propensity to remain in the current state is exploited; simply, asserting that if the temperature is cold for any given measurement, it is more likely than not that the subsequent measurement will also be made at a cold temperature. The same reasoning is also applied to normal temperature conditions (if only weather forecasting really were so simple!). These assumptions are reflected in the conditional probability distribution $P(H_{t+1}|H_t, d_t=0)$ shown in Table \ref{tab:CPD1Z24}. Here, the assumed distributions are sufficiently representative to demonstrate the risk-based active learning process. However, in practice, one may use degradation modelling, climate modelling and expert elicitation to develop these transition models. Moreover, it is worth acknowledging that, for some applications, it may be desirable to remove the effects of environmental and operational variables. This removal process has been demonstrated in \cite{Cross2011}.

\begin{table}[ht]
	\centering
	\caption{The conditional probability table $P(H_{t+1}|H_t, d_t)$ for $d_t = 0$.}
	\label{tab:CPD1Z24}       
	\begin{tabular}{c c| c c c c}
		
		\multicolumn{2}{c|}{\multirow{4}{*}{ }} & \multicolumn{4}{c}{$H_{t+1}$} \\
		
		&  & 1 & 2 & 3 & 4 \\
		\hline\noalign{\smallskip}
		\multirow{4}{*}{$H_t$}   & 1 & 0.7 & 0.28 & 0.015 & 0.005\\
		& 2  & 0.43  & 0.55 & 0.15 & 0.05\\
		& 3  & 0  & 0 & 0.8 & 0.2\\
		& 4  & 0  & 0 & 0 & 1\\
		\hline
	\end{tabular}
\end{table}

The counterpart to the conditional probability distribution shown in Table \ref{tab:CPD1Z24}, $P(H_{t+1}|H_t,d_t=1)$, specifying the state transition probabilities given $d_t = 1$ is shown in Table \ref{tab:CPD2Z24}. This conditional probability distribution is based on the assumption that the `perform maintenance' action returns the structure to one of the two undamaged states with high probability. The probabilities constituting this distribution, given the structure is in one of the two undamaged states initially, are specified such that the future undamaged state is independent of the action being undertaken. This assumption is, of course, a natural one to make due to the reasoning that (chaos theory aside) the act of repairing the bridge does not influence the weather. The remaining probabilities, conditional on the structure being in one of the two damaged states initially, are specified in a similar manner. Firstly, it is assumed that there is a small probability that the maintenance has no effect and the structure remains in its damaged state. Secondly, it is assumed that the remaining probability is attributed to each of the undamaged states in accordance with the stationary distribution obtained when considering the probability of being in normal or cold temperatures in the distant future. This assumption is made as the classes corresponding to damaged states do not distinguish between temperatures and therefore provide no information regarding which of the undamaged states the structure is likely to be returned to. More formally, it is asserted that,

\begin{multline}
P(H_{t+1}=1 \vee H_{t+1}=2|H_{t}=3 \vee H_{t}=4,d_t=1) = \\ P(H_{t+1}=1 \vee H_{t+1}=2|H_{t}=1 \vee H_{t}=2,d_t=1)^{\infty}
\end{multline}

\noindent
where $\vee$ is the OR logical operator. As it is implicitly assumed that the states $H_t$ are mutually exclusive this is also equivalent to the XOR logical operator.

\begin{table}[ht]
	\centering
	\caption{The conditional probability table $P(H_{t+1}|H_t, d_t)$ for $d_t = 1$.}
	\label{tab:CPD2Z24}       
	\begin{tabular}{c c| c c c c}
		
		\multicolumn{2}{c|}{\multirow{4}{*}{ }} & \multicolumn{4}{c}{$H_{t+1}$} \\
		
		&  & 1 & 2 & 3 & 4 \\
		\hline\noalign{\smallskip}
		\multirow{4}{*}{$H_t$}   & 1 & 0.7143 & 0.2857 & 0 & 0\\
		& 2  & 0.4388  & 0.5612 & 0 & 0\\
		& 3  & 0.5996  & 0.3904 & 0.01 & 0\\
		& 4  & 0.5996  & 0.3904 & 0 & 0.01\\
		\hline
	\end{tabular}
\end{table}

For simplicity, it is once again assumed that utilities may be attributed directly to the states of interest for this decision problem. The utility function $U(H_{t+1})$ is shown in Table \ref{tab:UHZ24}.  Here, the undamaged states associated with the bridge are assigned small positive utilities as reward for the bridge being functional with minimal risk of failure. The incipient damage state is assigned moderately-sized negative utility to reflect possible reduced functionality or low-to-moderate risk of failure. The advanced damage state is assigned a very large negative utility to reflect the high risk associated with the failure of the bridge.

\begin{table}[ht]
	\centering
	\caption{The utility function $U(H_{t+1})$.}
	\label{tab:UHZ24}       
	\begin{tabular}{c|c c c c}
		\hline\noalign{\smallskip}
		$H_{t+1}$ & $1$ & $2$ & $3$ & $4$\\
		\hline\noalign{\smallskip}
		$U(H_{t+1})$ & $10$ & $10$ & $-50$ & $-1000$\\
		\noalign{\smallskip}\hline
	\end{tabular}
\end{table}

Finally, the cost of inspection is specified to be $C_{\text{ins}} = 30$. This moderate cost reflects the time required to inspect a large-scale structure such as a bridge. 

\subsection{Statistical classifier}

Once again, a probabilistic Gaussian mixture model was used as the statistical classifier; trained repeatedly in a supervised manner as $\mathcal{D}_l$ was extended via risk-based querying process. As previously mentioned, the discriminative features used were the first four natural frequencies of the bridge, normalised with respect to the mean and standard deviations. The targets of the classifier were the four states of interest corresponding to two undamaged and two damaged states. 

\subsection{Results}

\begin{figure}[htbp!]
	\centering
	\scalebox{0.4}{
		\includegraphics{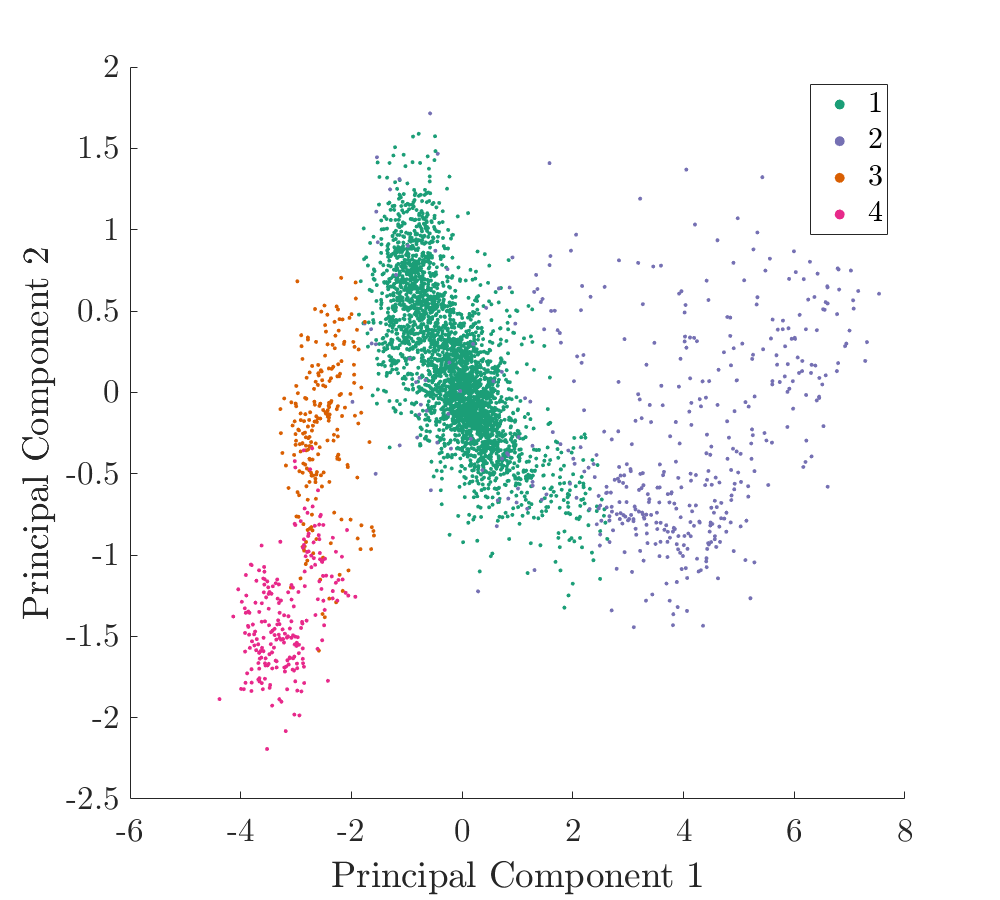}
	}
	\caption{A two-dimensional projection of the feature space for the complete Z24 dataset.}
	\label{fig:AllDataZ24}
\end{figure}

As previously mentioned, the Z24 dataset comprises of 3932 observations. Once again, these data were divided in half to form a training dataset and a test dataset. A small (1\%) random subset of the training dataset $\mathcal{D}$ was assigned to the initial labelled dataset $\mathcal{D}_l$ and the remaining data assigned to the unlabelled dataset $\mathcal{D}_u$ to be sequentially presented to the decision model in the risk-based active learning process. To facilitate visualisation, the four-dimensional feature space was projected down onto two-dimensions using \textit{principal component analysis} \cite{Jolliffe2016}. The two-dimensional projection of the Z24 dataset is shown in Figure \ref{fig:AllDataZ24}.

\begin{figure}[htbp!]
	\centering
	\scalebox{0.4}{
		\includegraphics{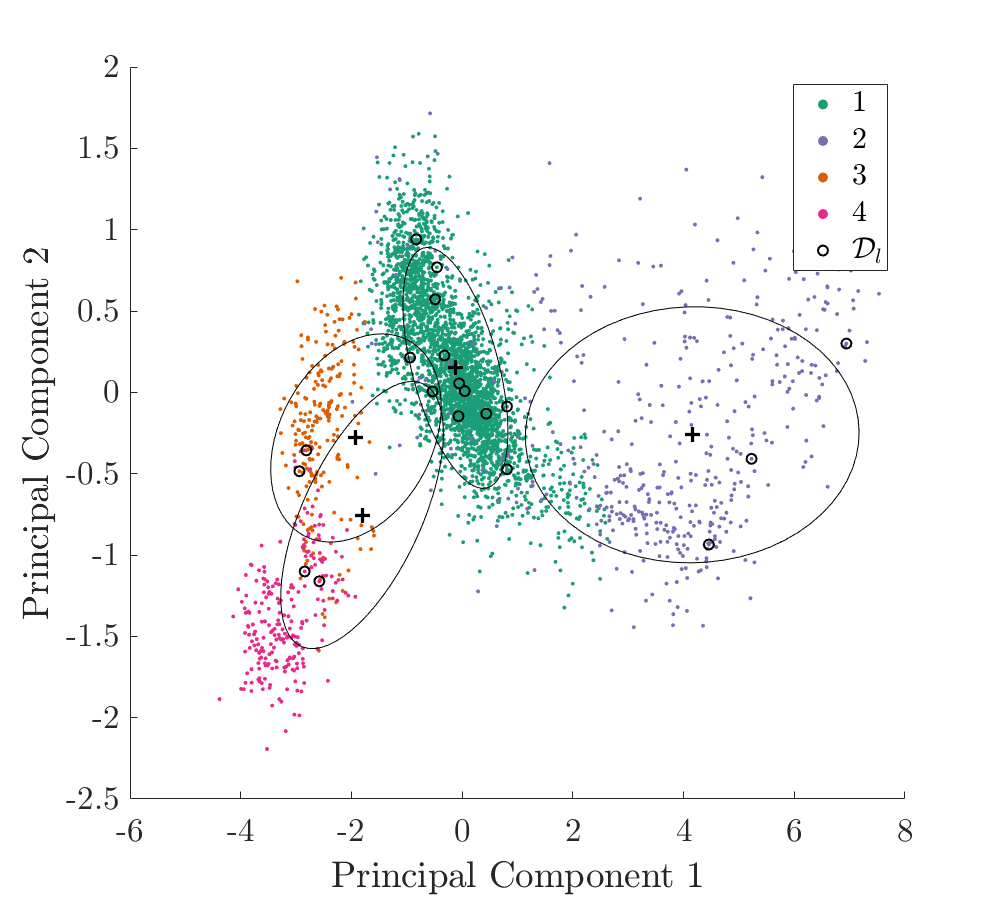}
	}
	\caption{A two-dimensional projection of an initial statistical classifier $p(\bm{\nu}_t, H_t, \bm{\Theta})$ learned from the initial labelled dataset $\mathcal{D}_l$; \textit{maximum a posteriori} (MAP) estimate of the mean (\textbf{+}) and covariance (ellipses represent $2\sigma$).}
	\label{fig:Z24init}
\end{figure}

An example of an initial model learned from the subset $\mathcal{D}_l$ is shown in Figure \ref{fig:Z24init} with data in $\mathcal{D}_l$ circled.  It can be seen from Figure \ref{fig:Z24init}, that it appears the Gaussian distribution best learned corresponds to the class $H_t = 1$. The reasons for this are twofold, the data for this class are positioned closest to zero-mean of the prior, and more datapoints belonging to this class were, by chance, included in the initial labelled dataset $\mathcal{D}_l$. The other three distributions appear to have been learned poorly, being heavily influenced by the prior.

\begin{figure}[htbp!]
	\centering
	\scalebox{0.4}{
		\includegraphics{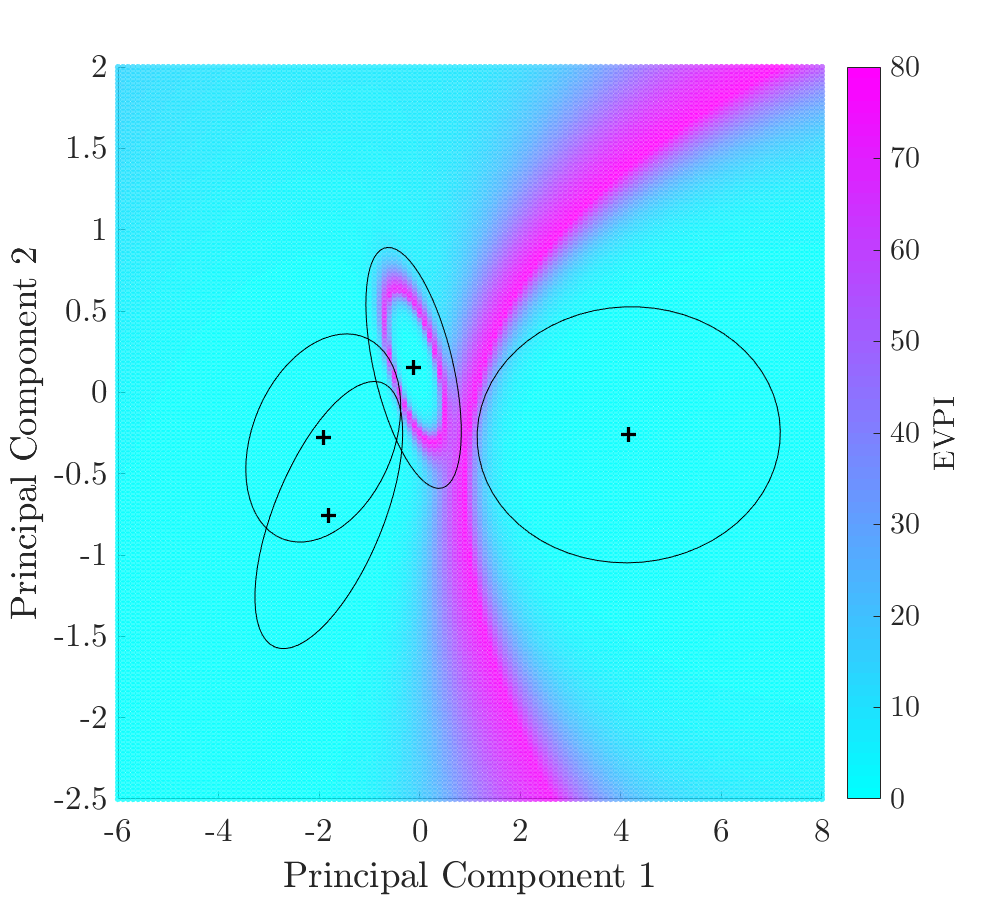}
	}
	\caption{The EVPI over the two-dimensional projection the feature space given the two-dimensional projection of the initial model shown in Figure \ref{fig:Z24init}.}
	\label{fig:Z24initEVPI}
\end{figure}

Figure \ref{fig:Z24initEVPI} shows the EVPI contours over the projected feature space. It can be seen that there are regions of high expected value around the edges of the normal undamaged cluster (Class 1), this is intuitively understood by considering the adjacent regions of low expected value. The low-value region bounded by the high-value region occurs as a result of the classifier being confident that the structure is in its undamaged condition and therefore the decision-maker is confident that the optimal decision is `do nothing'. Conversely, the low-value region enclosing the high-value region occurs as a result of the tolerable risk of damage/failure having been exceeded, as such the decision-maker is confident that the optimal decision for data in this region is `perform maintenance'. Again, Figure \ref{fig:Z24initEVPI} shows that there are larger swathes of high-value, corresponding to regions of high uncertainty between undamaged and damaged states.
	
\begin{figure}[htbp!]
	\centering
	\scalebox{0.4}{
		\includegraphics{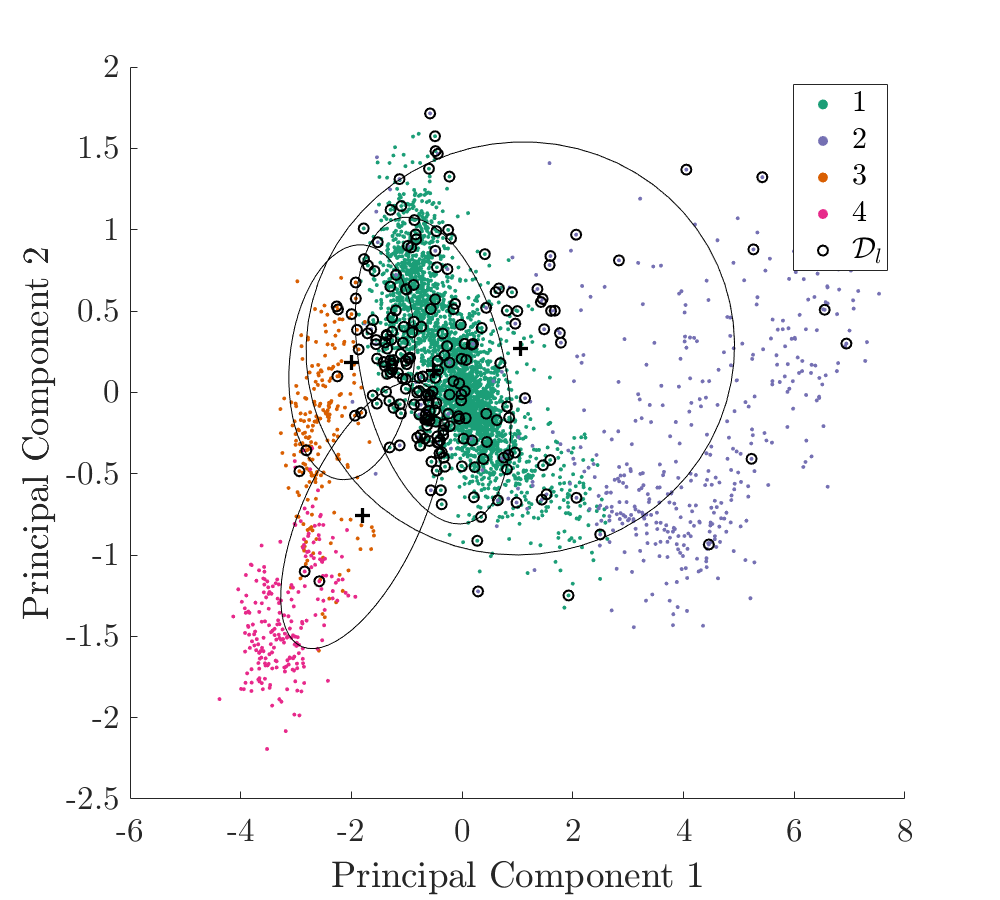}
	}
	\caption{A two-dimensional projection of a final model statistical classifier $p(\bm{\nu}_t, H_t, \bm{\Theta})$ learned from the extended labelled dataset $\mathcal{D}_l$; \textit{maximum a posteriori} (MAP) estimate of the mean (\textbf{+}) and covariance (ellipses represent $2\sigma$).}
	\label{fig:Z24final}
\end{figure}

The laballed dataset $\mathcal{D}_l$ was extended via the risk-based active learning process as data were presented to the decision process in sequential order one-at-a-time. Again, labels for datapoints were queried and the classifier retrained on the extended dataset when the criterion $\text{EVPI} > C_{\text{ins}}$ was satisfied. The final model, corresponding to the updated version of the initial model shown in Figure \ref{fig:Z24init} following risk-based active learning, is shown in Figure \ref{fig:Z24final}.

It can be seen in Figure \ref{fig:Z24final} that the active learner has preferentially queried datapoints belonging to Class 1 and, to a lesser extent, Class 2. Data on the boundary between Classes 1 and 3 appears to have been particularly heavily sampled. This observation is, again, to be expected when considering the distribution of EVPI shown in Figure \ref{fig:Z24initEVPI}, which shows low-value regions in the vicinity of advanced-damage data and high-value regions on the boundary of Class 1. 

\begin{figure}[htbp!]
	\centering
	\scalebox{0.4}{
		\includegraphics{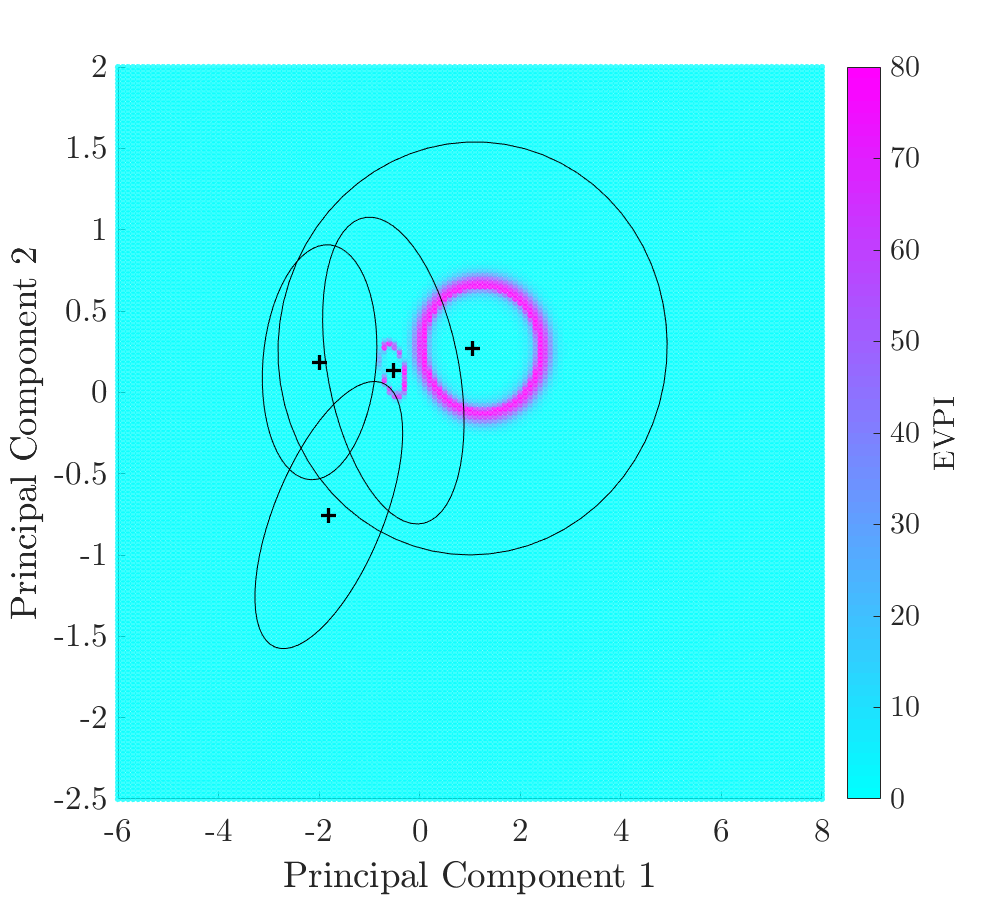}
	}
	\caption{The EVPI over the two-dimensional projection the feature space given a 2-dimensional projection of the final model shown in Figure \ref{fig:Z24final}.}
	\label{fig:Z24finalEVPI}
\end{figure}

Figure \ref{fig:Z24finalEVPI} shows the EVPI over the projected feature space given a projection of the final model. Figure \ref{fig:Z24finalEVPI} shows very well-defined `rings' of high expected value. Once again, the areas inside these rings can be considered regions on the feature space where the classifier is sufficiently confident that the structure is currently, and will be in the subsequent time-step, in an undamaged state such that the decision-maker can be confident that `do nothing' is the optimal action without the need for inspection of the structure. Likewise, areas outside of the rings correspond to regions of the feature space where the decision-maker may be confident that `perform maintenance' is the optimal action.

\begin{figure}[htbp!]
	\centering
	\scalebox{0.4}{
		\includegraphics{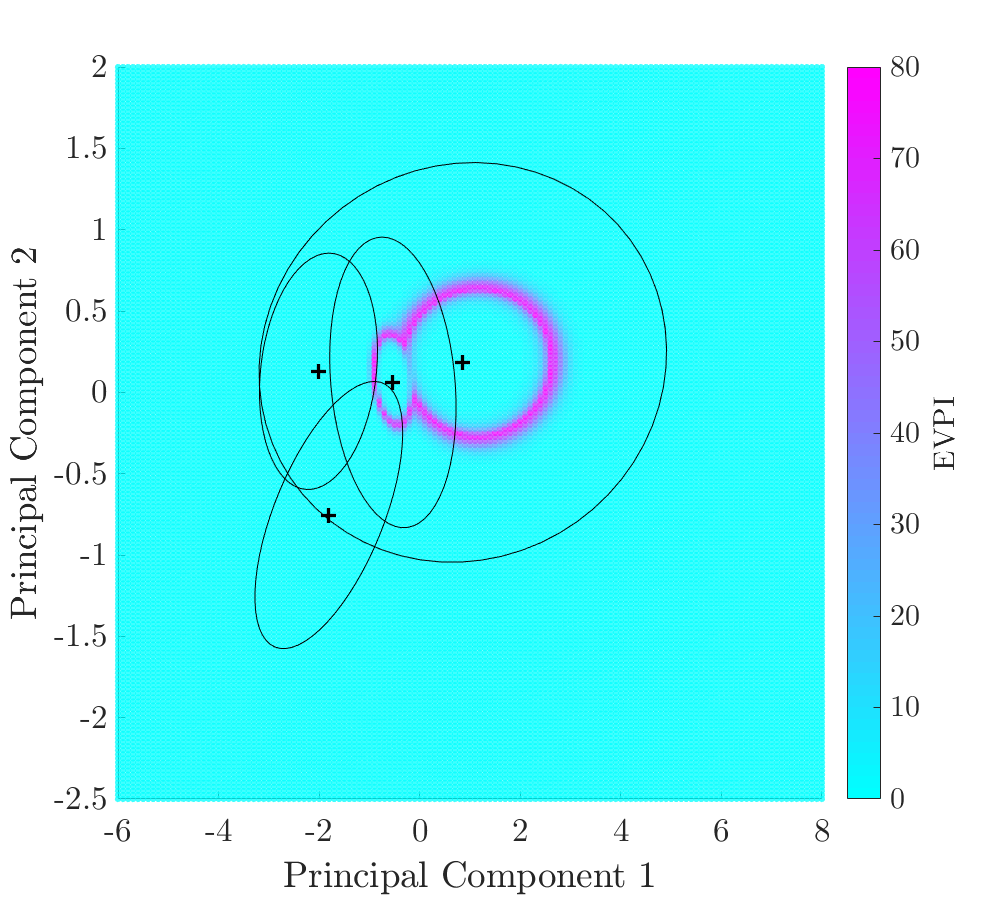}
	}
	\caption{The EVPI over the two-dimensional projection the feature space given a 2-dimensional projection of the final model learned when $U(H_t=4)=-500$.}
	\label{fig:Z24finalEVPI2}
\end{figure}

By comparison of Figures \ref{fig:Z24initEVPI} and \ref{fig:Z24finalEVPI}, it can be seen that the swathes of high-value feature space are replaced with a ring of high-value, associated with the cold temperature undamaged class. Additionally, the ring associated with the normal undamaged class becomes tighter. These phenomena indicate that, with hindsight, one can deem an agent utilising the initial classifier to be over-confident in its decision-making. For the current case study, it is to be expected that the areas in which a decision-maker can be confident that the `do nothing' action are small because of the large negative utility associated with the advanced damage class and the large resulting risk. In fact, if one reduces the cost associated with the bridge being in its advanced damage state, then one can expect the rings to grow larger as the region of tolerable risk expands. This can be observed in the merging of the rings shown in Figure  \ref{fig:Z24finalEVPI2}. The final model shown in Figure \ref{fig:Z24finalEVPI2} was learned with the cost associated with the advanced damage state reduced from 1000 to 500.

\begin{figure}[htbp!]
	\centering
	\scalebox{0.4}{
		\includegraphics{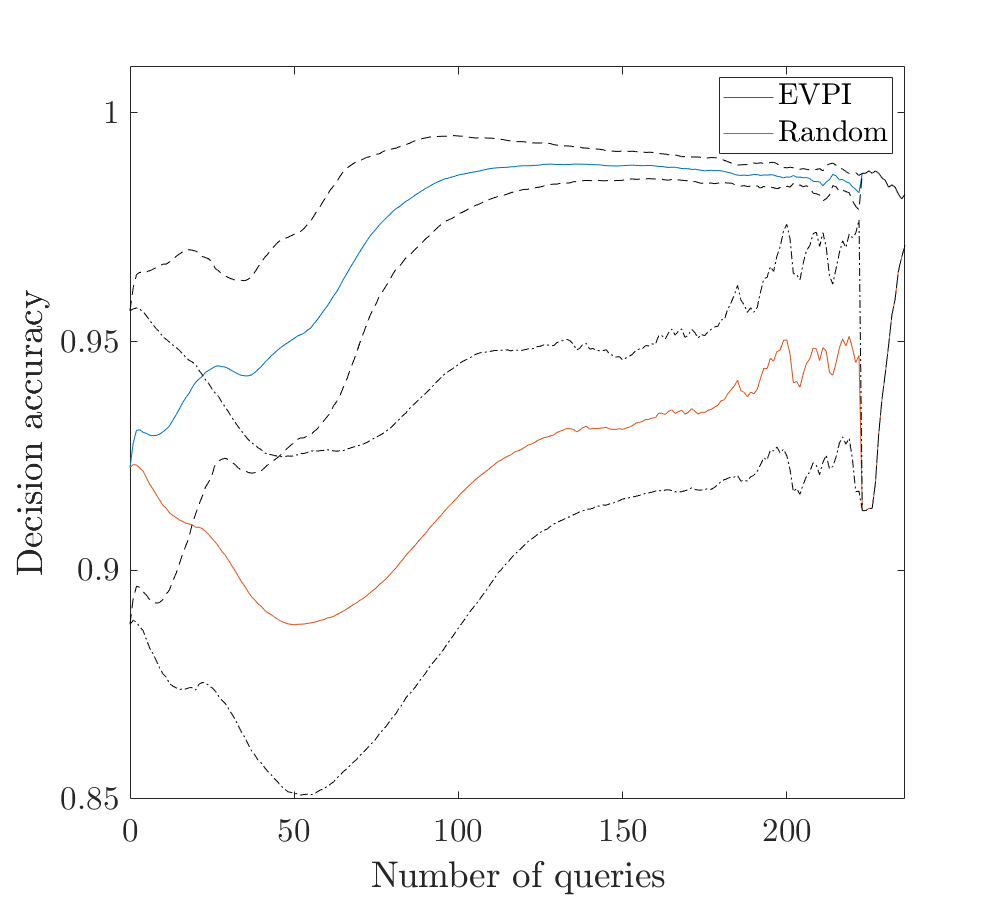}
	}
	\caption{The variation in decision accuracy with number of label queries for an agent utilising a statistical classifier trained on $\mathcal{D}_l$ extended via (i) risk-based active querying (EVPI) and (ii) random sampling (Random). The dashed lines show $\pm 1\sigma$.}
	\label{fig:Z24DecAcc}
\end{figure}

Throughout the active learning process, the decision-making performance of an agent was evaluated by evaluating the decision accuracy for data in the independent test set. The active learning process was repeated 1000 times with differing random number generator seeds such that the data in $\mathcal{D}$ and the initial subset $\mathcal{D}_l$ were randomly varied. The mean and standard deviation of the decision accuracy as a function of the number of queries is shown in Figure \ref{fig:Z24DecAcc}. For comparison, the decision accuracy for a classifier training on a labelled dataset extended via random sampling is also shown.

The decision accuracy can be seen to improve almost monotonically when using risk-based active learning, as opposed to random sampling which initially results in a degradation in performance before improving. For risk-based active learning, in addition to the mean value of decision accuracy reaching close to unity with fewer queries than random sampling, the variance of the decision accuracy also converges more rapidly. This result indicates that using risk-based active learning may result in more consistent decision-making performance. Upon closer examination, it can be seen that, for higher numbers of queries, the decision accuracy slightly declines. This observation and the initial decrease in accuracy for random sampling can be explained by \textit{sampling bias}, which is discussed further in the penultimate Section \ref{sec:discussion}.

To summarise, risk-based active learning was applied to a `real-world' case study, specifically the Z24 Bridge benchmark for SHM. A simple maintenance decision problem was constructed for the bridge. A probabilistic Gaussian mixture model was employed to distinguish between four salient states of interest using natural frequencies as a discriminative feature. The EVPI of incipient data points with respect to the decision process was used to guide the querying of health-state labels where querying would correspond to the inspection of the bridge.

\section{Discussion}
\label{sec:discussion}

The results presented in Sections \ref{sec:example} and \ref{sec:Z24} indicate that making inspections of structures according to a risk-based active learning heuristic may provide a cost-effective method of developing statistical classifiers for use in decision processes in situations when no, or limited, labelled data are available \textit{a priori}. Moreover, it is interesting to note from Figures \ref{fig:FinalModel} and \ref{fig:Z24final} that classifiers do not necessarily need to accurately model the entire feature space to be useful in decision processes, and in fact, well-fitting models can be foregone in favour of decision performance. That being said, a statistical model that is able to accurately predict health-state labels has the potential to provide additional information such as damage locations and types that may be useful in directing more specific types of maintenance and coordinating repair teams - albeit at a potentially higher cost. Indeed, as the dimensionality of a decision space increases, the need for high classification accuracy becomes more crucial.  Nevertheless, it can be said with confidence that the appropriate machine learning paradigm to be used for statistical classifier development in SHM is highly dependent on the context in which the monitoring system is being employed; taking into account factors such as: data availability, knowledge of relevant physics, and the decision support application of the monitoring system itself.  

The risk-based approach to active learning presented in the current paper is not without its limitations. Notably, in order to evaluate EVPI as presented, it is necessary to assume that the number of classes (i.e.\ health states of interest) are known prior to the implementation of a monitoring system. This limits the flexibility of the classifier and may result in the mistreatment of unforeseen health states and associated failure events within the decision framework. Additionally, it is assumed that perfect information of the health state can be acquired my means of inspection. During the inspection procedure, human error may be introduced, and in some scenarios it may be vital to account for this uncertainty in the active learning framework by relaxing the perfect information assumption when possible.

\subsection{Sampling bias}

A noteworthy observation from Sections \ref{sec:example} and \ref{sec:Z24} that bears further discussion is that, whilst decision accuracy may be increased via risk-based active learning, after a certain number of queries have been made, any incipient data points whose true labels may have high expected value with respect to a structural maintenance decision process, may, in fact, be detrimental to the performance of a decision-maker when incorporated into the statistical classifier on which it relies.

A likely explanation for this observation is the phenomenon known as \textit{sampling bias}; a known issue associated with active learning that has been documented to impact upon classification performance \cite{Bull2019,Dasgupta2008}. Sampling bias occurs when specific regions of feature-space are over/under-sampled resulting in a training dataset that is not representative of the underlying distributions; this can clearly be observed in Figure \ref{fig:FinalModel}. In the numerical example presented in Section \ref{sec:example}, sampling bias manifests as unrepresentative mixing proportions $\bm{\lambda}$, which may result in overconfident misclassifications that subsequently cause erroneous actions to be decided.

\subsection{Future work}

To overcome the dangers of sampling bias, two potential solutions could be considered for future work. The first would be to establish a heuristic-based methodology for switching between value of information-based and uncertainty-based measures for guided sampling. This would have the effect of establishing decision boundaries by querying regions of the feature space with high value of information, while also exploring low likelihood and high information (including low-value) regions of the feature space. An alternative approach would be to incorporate semi-supervised learning techniques \cite{Chapelle2006,Bull2018}, such that the label predictions for the unlabelled data $\mathcal{D}_u$ may be utilised to retain representative estimates for the mixing proportions.

Another interesting avenue for further investigation is that of risk-based active learning applied to more complex decision spaces, such that the requirements of the statistical classifiers are extended to damage localisation and the identification of different types of damage. The case study presented in the current paper considered only a simply binary decision between `do nothing' and `perform maintenance'. By increasing the number of decisions, and therefore the number of decision boundaries to be learned, the generalisation capabilities of the framework will be tested, as no doubt more queries will be required. This investigation will potentially result in the decision accuracy of the value-based querying approach being superseded by that of the random sampling approach for higher numbers of queries.

\section{Conclusions}
\label{sec:conclusions}

The aim of the current paper has been to present a risk-based active learning approach for SHM. The approach utilises the expected value of perfect information of incipient data points to instigate inspections, such that structural health information may be obtained and incorporated into probabilistic classifiers. The methodology was demonstrated on a numerical case study to aid in visualisation and understanding of the risk-based active learning process. Additionally, the approach was demonstrated on an experimental dataset obtained from a previously-existing bridge, thereby highlighting its potential applicability to `real-world' engineering problems. The results of the case studies indicated that the risk-based approach to active learning has the potential to provide a cost-effective solution to the development of decision-supporting SHM systems. This finding is valuable as, ordinarily, the comprehensive labelled datasets necessary for the fully-supervised learning of statistical classifiers are seldom available at the inception of a monitoring system.

	\section*{Acknowledgements}
		The authors would like to acknowledge the support of the UK EPSRC via the Programme Grant EP/R006768/1. KW would also like to acknowledge support via the EPSRC Established Career Fellowship EP/R003625/1.
	
	%
	\section*{Conflict of interest}
	
	The authors declare that they have no conflict of interest.

	\bibliographystyle{elsarticle-num}
	\bibliography{RBAL_JP}

\end{document}